\documentclass{article}

\usepackage{microtype}
\usepackage{graphicx}
\usepackage{subfigure}
\usepackage{booktabs} 
\usepackage{amsmath}
\usepackage{hyperref}
\usepackage{amssymb}
\usepackage{amsthm}
\usepackage{url}
\usepackage{subfigure}
\usepackage{booktabs}
\usepackage{multirow}
\usepackage{pdfpages}
\newcommand{\blambda}{\boldsymbol\lambda}
\usepackage{siunitx}
\newtheorem{theorem}{Theorem}

\newtheorem{remark}{Remark}

\newtheorem{definition}{Definition}

\usepackage{hyperref}

\usepackage[accepted]{icml2019}

\icmltitlerunning{Breaking the 
Limits of Message Passing Graph Neural Networks}

\begin{document}

\twocolumn[
\icmltitle{Breaking the Limits of Message Passing Graph Neural Networks}




\begin{icmlauthorlist}
\icmlauthor{Muhammet Balcilar}{to,inter}
\icmlauthor{Pierre~H\'eroux}{to}
\icmlauthor{Benoit~Ga\"uz\`ere}{insa}
\icmlauthor{Pascal~Vasseur}{to,mibs}
\icmlauthor{S\'ebastien~Adam}{to}
\icmlauthor{Paul~Honeine}{to}
\end{icmlauthorlist}

\icmlaffiliation{to}{LITIS Lab, University of Rouen Normandy, France}
\icmlaffiliation{inter}{InterDigital, France}
\icmlaffiliation{insa}{LITIS Lab, INSA Rouen Normandy, France}
\icmlaffiliation{mibs}{MIS Lab, Université de Picardie Jules Verne, France}
\icmlcorrespondingauthor{Muhammet Balcilar}{muhammetbalcilar@gmail.com}

\icmlkeywords{Graph Neural Networks, Expressive Power, 1-WL test, Spectral Design}

\vskip 0.3in
]



\printAffiliationsAndNotice{} 

\begin{abstract}

Since the Message Passing (Graph) Neural Networks (MPNNs) have a linear complexity with respect to the number of nodes when applied to sparse graphs, they have been widely implemented and still raise a lot of interest even though their theoretical expressive power is limited to the first order Weisfeiler-Lehman test (1-WL). In this paper, we show that if the graph convolution supports are designed in spectral-domain by a non-linear custom function of eigenvalues and masked with an arbitrary large receptive field, the MPNN is theoretically more powerful than the 1-WL test and experimentally as powerful as a 3-WL existing models, while remaining spatially localized. Moreover, by designing custom filter functions, outputs can have various frequency components that allow the convolution process to learn different relationships between a given input graph signal and its associated properties.
So far, the best 3-WL equivalent graph neural networks have a computational complexity in $\mathcal{O}(n^3)$ with memory usage in $\mathcal{O}(n^2)$, consider non-local update mechanism and do not provide the spectral richness of output profile. The proposed method overcomes all these aforementioned problems and reaches state-of-the-art results in many downstream tasks.  

\end{abstract}

\section{Introduction}

In the past few years, finding the best inductive bias for relational data represented as graphs has gained a lot of interest in the machine learning community. Node-based message passing mechanisms relying on the graph structure have given rise to the first generation of Graph Neural Networks (GNNs) called Message Passing Neural Networks (MPNNs) \cite{gilmer17:_neural_messag_passin_quant_chemis}. These algorithms spread each node features to the neighborhood nodes using trainable weights. These weights can be shared with respect to the distance between nodes (Chebnet GNN) \cite{defferrard16:_convol_neural_networ_graph_fast}, to the connected nodes features (GAT for graph attention network) \cite{velivckovic2017graph} and/or to edge features \cite{bresson2017residual}. When considering sparse graphs, the memory and computational complexity of such approaches are linear with respect to the number of nodes. As a consequence, these algorithms are feasible for large sparse graphs 
and thus have been applied with success on many downstream tasks \cite{dwivedi2020benchmarking}. 


Despite these successes and these interesting computational properties, it has been shown that MPNNs are not powerful enough \cite{xu2018how}. Considering two non-isomorphic graphs that are not distinguishable by the first order Weisfeiler-Lehman test (known as the 1-WL test), existing maximum powerful MPNNs embed them to the same point. Thus, from a theoretical expressive power point of view, these algorithms are not more powerful than the 1-WL test. Beyond the graph isomorphism issue, it has also been shown that many other combinatorial problems on graph cannot be solved by MPNNs \cite{satonips}.  

In \cite{universalmaron19a,NIPS2019_8931}, it 
has been proven that in order to reach universal approximation, higher order relations are 
required. In this context, some powerful models that are equivalent to the 3-WL test were proposed. For instance, \cite{maron2019provably} proposed the model PPGN (Provably Powerful Graph Network) that mimics the second order Folklore WL test (2-FWL), which is equivalent to the 3-WL test. In \cite{123lehmango}, they proposed 
to use message passing between 1, 2 and 3 order node tuples hierarchically, thus reaching the 3-WL expressive power. However, using such relations makes both memory usage and computational complexities grown exponentially. Thus, it is not feasible to have universal approximation models in practice. 

In order to increase the theoretical expressive power of MPNNs by keeping the linear complexity mentioned above, some researchers proposed to partly randomize node features  \cite{abboud2020surprising,sato2020random} or to add a unique label
~\cite{murphy19a} in order to have the ability to distinguish two non-isomorphic graphs that are not distinguished by the 1-WL test. These solutions need massively training samples and involve slow convergence.  \cite{bouritsas2020improving,Dasoulascoloring} proposed to use a preprocessing step to extract some features that cannot be extracted by MPNNs. Thus, the expressive power of their GNN is improved. However, these handcrafted features need domain expertise and a feature selection process among an infinite number of possibilities.  

All these studies target more theoretically powerful models, closer to universal approximation. However, this does not always induce a better generalization ability.
Since most of the realistic problems are given with many node/edge features (which can be either continuous or discrete), there is almost no pair of graphs that are not distinguishable by the 1-WL test in practice. In addition, theoretically more powerful methods use non-local updates, breaking one of the most important inductive bias in Euclidean learning named locality principle \cite{battaglia2018relational}. These may explain why theoretical powerful methods cannot outperform MPNNs on many downstream tasks, as reported in \cite{dwivedi2020benchmarking}. On the other hand, it is obvious that 1-WL equivalent GNNs are not expressive enough since they are not able to 
count 
some simple structural features such as 
cycles or triangles~\cite{arvind2020weisfeiler,chen2020can,bouritsas2020improving,vignac2020building}, which are informative for some social or chemical graphs. 
Finally, another important aspect mentioned by a recent paper \cite{balcilar2021analyzing} concerns the spectral ability of GNN models. It is shown that a vast majority of the MPNNs actually work as low-pass filters, thus reducing their expressive power. 

In this paper, we propose to design graph convolution in the spectral domain with custom non-linear functions of eigenvalues and by masking the convolution support with desired length of receptive field. In this way, we have (i) a spatially local updates process, (ii) linear memory and computational complexities (except the eigendecomposition in preprocessing step), (iii) enough spectral ability and (iv) a model that is theoretically more powerful than the 1-WL test, and experimentally as powerful as PPGN. 
Experiments show that the proposed model can distinguish pairs of graphs that cannot be distinguished by 1-WL equivalent MPNNs. It is also able to count some substructures that 1-WL equivalent MPNNs cannot. Its spectral ability enables to produce various kind of spectral components in the output, while the vast majority of the GNNs including higher order WL equivalent models do not. Finally, thanks to the sparse matrix multiplication, it has linear time complexity except the eigendecomposition in preprocessing step.


The paper is structured as follows. In Section 2, we set the notations and the general framework used in the following. Section 3 is dedicated to the characterization of WL test, which is the backbone of our theoretical analysis. It is followed by our findings in Section 4 on analysing the expressive power of MPNNs and our solutions to improve expressive power of MPNNs in Section 5. The experimental results and conclusion are the last two section of this paper.

\section{Generalization of Spectral and Spatial MPNN}

Let $G$ be a graph with $n$ nodes and an arbitrary number of edges. Connectivity is given by the adjacency matrix $A \in \{0,1\}^{n \times n}$ and features are defined on nodes by $X \in \mathbb{R}^{n \times f_0}$, with $f_0$ the length of feature vectors. For any matrix $X$, we used $X_i$, $X_{:j}$ and $X_{i,j}$ to refer to its $i$-th column vector, $j$-th row vector and ($i,j$)-th entry,  respectively. A graph Laplacian is given by $L=D-A$ (or $L=I-D^{-1/2}AD^{-1/2}$) where $D \in \mathbb{R}^{n \times n}$ is the diagonal degree matrix and $I$ is the identity. Through an eigendecomposition, $L$ can be written by $L=U {diag}(\blambda) U^T$ where each column of $U \in \mathbb{R}^{n \times n}$ is an eigenvector of $L$, $\blambda \in \mathbb{R}^n$ gathers the eigenvalues of L and ${diag}(\cdot)$ creates a diagonal matrix whose diagonal elements are from a given vector. We use superscripts to refer to vectors or matrices evolving through iterations or layers. For instance, $H^{(l)} \in \mathbb{R}^{n \times f_l}$ refers to the node representation on layer $l$ whose feature dimension is $f_l$.

GNN models rely on a set of layers where each layer takes the node representation of the previous layer $H^{(l-1)}$ as input and produces a new representation $H^{(l)}$, with $H^{(0)}=X$. According to the domain which is considered to design the layer computations, GNNs are generally classified as either spectral or spatial \citep{wu19:_compr_survey_graph_neural_networ,chami2020machine}. Spectral GNNs rely on the spectral graph theory~\citep{OL-CHUNG-1997}. In this framework, signals on graphs are filtered using the eigendecomposition of the graph Laplacian \citep{shuman2013emerging}. By transposing the convolution theorem to graphs, the spectral filtering in the frequency domain can be defined by $x_{flt} =  U {diag}( \Omega(\blambda)) U^\top x$, where $\Omega(.)$ is the desired filter function which needs to be learnt by back-propagation. On the other hand, spatial GNNs, such as GCN (graph convolutional network) \cite{kipf16:_semi} and GraphSage \cite{hamilton2017sage}, consider two operators, one that aggregates the connected nodes messages and one that updates the concerned node representation.

In a recent paper \cite{balcilar2021analyzing}, it was explicitly shown that both spatial and spectral GNNs are MPNN, taking the general form
\begin{equation}
   \label{eq:mpgnn}
   H^{(l+1)} = \sigma \Big( \sum_s C^{(s)} H^{(l)} W^{(l,s)} \Big),
\end{equation}
where $C^{(s)} \in \mathbb{R}^{n \times n}$ is the $s$-th convolution support that defines how the node features are propagated to the neighboring nodes and $W^{(l,s)} \in \mathbb{R}^{f_l \times f_{l+1}}$ is the trainable matrix for the $l$-th layer and $s$-th support. 
Within this generalization, GNNs differ from each other by the design of the convolution supports $C^{(s)}$. If the supports are designed in the spectral domain by $\Phi_s(\blambda)$, the convolution support needs to be written as $C^{(s)} = U {diag} (\Phi_s(\blambda)) U^\top$. 

One can see that as long as $C^{(s)}$ matrices are sparse (number of edges is defined by some constant multiplied by the number of nodes), MPNN in Eq.\ref{eq:mpgnn} has linear memory and computational complexities with respect to the number of nodes. Because, the valid entries in $C^{(s)}$ that we need to keep is linear with respect to the number of nodes and thank to the sparse matrix multiplication $C^{(s)}H^{(l)}$ takes linear time with respect to the number of edges thus nodes as well. 

\section{Characterization of Weisfeiler-Lehman} 
\label{section:WL}
The universality of a GNN is based on its ability to embed two non-isomorphic graphs to distinct points in the target feature space. A model that can distinguish all pairs of non-isomorphic graphs is a universal approximator. Since the graph isomorphism problem is NP-intermediate \cite{takapoui2016linear}, the Weisfeiler-Lehman Test (abbreviated WL-test), which gives sufficient but not enough evidence of graph isomorphism, is frequently used for characterizing GNN expressive power. The classical vertex coloring WL test can be extended by taking into account higher order of node tuple within the iterative process. These extensions are denoted as $k$-WL test, where $k$ is equals to the order of the tuple. These tests are described in Appendix \ref{section:app_wl}. 

It is shown in \cite{arvind2020weisfeiler} that for $k\geq2$, $(k+1)$-WL $>$ $(k)$-WL, i.e., higher order of tuple leads to a better ability to distinguish two non-isomorphic graphs. For $k=1$, this statement is not true, and 2-WL is not more powerful than 1-WL \cite{maron2019provably}. To clarify this point, the
Folkore WL (FWL) test has been defined such that 1-WL=1-FWL, but for $k \geq 2$, we have $(k+1)$-WL $\approx$ $(k)$-FWL~\cite{maron2019provably}.

In literature, some confusions occur among the two versions. Some papers use WL test order \cite{123lehmango, maron2019provably}, while others use FWL order under the name of WL such as in \cite{abboud2020surprising,arvind2020weisfeiler,takapoui2016linear}. In this paper, we explicitly mention both WL and FWL equivalent.

In order to better understand the capability of WL tests, some papers attempt to characterize these tests using a first order logic \cite{immerman1990describing,barcelo2019logical}. Consider two unlabeled and undirected graphs represented by their adjacency matrices $A_G$ and $A_H$. These two graphs are said $k$-WL (or $k$-FWL) equivalent, and denoted $A_G\equiv_{k-WL}A_H$, if they are 
indistinguishable by a $k$-WL (or $k$-FWL) test.




Recently \cite{brijder2019expressive,geerts2020expressive} proposed a new Matrix Language called MATLANG. This language includes different operations on matrices and makes some explicit connections between specific dictionaries of operations and the 1-WL and 3-WL tests. Expressive power varies with the operations included in each dictionnary.

\begin{definition}
$ML(\mathcal{L})$ is a matrix language with an allowed operation set $\mathcal{L}=\{op_1,\dots op_n\}$, where $op_i \in \{.,+,^\top,diag,tr,{\bf1},\odot, \times,f\}$. The possible operations are matrices multiplication and addition, matrix transpose, vector diagonalization, matrix trace computation, column vector full of 1, element-wise matrix multiplication, matrix/scalar multiplication and element-wise custom function operating on scalars or vectors.
\end{definition}

\begin{definition}
$e(X) \in \mathbb{R}$ is a sentence in $ML(\mathcal{L})$ if it consists of any possible consecutive operations in $\mathcal{L}$, operating on a given matrix $X$ and resulting in a scalar value. 
\end{definition}

As an example, $e(X)={\bf1}^{\top}X^2{\bf1}$ is a sentence of $ML(\mathcal{L})$ with $\mathcal{L}=\{.,^\top,{\bf1}\}$, computing the sum of all elements of square matrix $X$. In the following, we are interested in languages $\mathcal{L}_1,\mathcal{L}_2$ and $\mathcal{L}_3$ that have been used for characterizing the WL-test in \cite{geerts2020expressive}. These results are given next.

\begin{remark}
 \label{rm:rm2}
Two adjacency matrices are indistinguishable by the 1-WL test if and only if $e(A_G) = e(A_H)$ for all $e \in \mathcal{L}_1$  
with $\mathcal{L}_1=\{.,^\top,{\bf1},diag\}$. Hence, all possible sentences in $\mathcal{L}_1$ are the same for 1-WL equivalent adjacency matrices. Thus, $A_G\equiv_{1-WL}A_H \leftrightarrow A_G\equiv_{ML(\mathcal{L}_1)}A_H$. (see Theorem 7.1 in \cite{geerts2020expressive})
\end{remark}

\begin{remark}
 \label{rm:rm3}
$ML(\mathcal{L}_2)$ with $\mathcal{L}_2=\{.,^\top,{\bf1},diag,tr\}$ is strictly more powerful than $\mathcal{L}_1$, i.e., than the 1-WL test, but less powerful than the 3-WL test. (see Theorem 7.2 and Example 7.3 in  \cite{geerts2020expressive})
\end{remark}

\begin{remark}
 \label{rm:rm5}
Two adjacency matrices are indistinguishable by the 3-WL test if and only if they are indistinguishable by any sentence in $ML(\mathcal{L}_3)$ with  $\mathcal{L}_3=\{.,^\top,{\bf1},diag,tr,\odot\}$. Thus, $A_G\equiv_{3-WL}A_H \leftrightarrow A_G\equiv_{ML(\mathcal{L}_3)}A_H$. (see Theorem 9.2 in \cite{geerts2020expressive})
\end{remark}

\begin{remark}
 \label{rm:rm4}
Enriching the operation set to $\mathcal{L}^+=\mathcal{L} \cup \{ +,\times,f\}$ where $\mathcal{L} \in (\mathcal{L}_1,\mathcal{L}_2,\mathcal{L}_3$) does not improve the expressive power of the language. Thus, $A_G\equiv_{ML(\mathcal{L})}A_H \leftrightarrow A_G\equiv_{ML(\mathcal{L}^+)}A_H$. (see Proposition~7.5 in \cite{geerts2020expressive})
\end{remark}

\section{How Powerful are MPNNs?}

This section presents some results about the theoretical expressive power of state-of-the-art MPNNs. Those results are derived using the MATLANG language  \cite{geerts2020expressive} and more precisely the remarks of the preceding section. Proofs of the theorems are given in Appendix \ref{section:app_prof}.  

\begin{theorem}
\label{th:thmpnn}
MPNNs such as GCN, GAT, GraphSage, GIN (defined in Appendix~\ref{section:implementation}) cannot go further than operations in $\mathcal{L}_1^+$. Thus, they are not more powerful than the 1-WL test.
\end{theorem}

This result has already been given in \cite{xu2018how}, which proposed GIN-$\epsilon$ (GIN for Graph Isomorphism Network) and showed that it is the unique MPNN which is provably exact the same powerful with the 1-WL test, 
while the rest of MPNNs are known to be less powerful than 1-WL test.

Chebnet is also known to be  not more powerful than the 1-WL test. However, the next theorem states that it is true if the maximum eigenvalues are the same for both graphs. For a pair of graphs whose maximum eigenvalues are not equal, Chebnet is strictly more powerful than the 1-WL test. 

\begin{theorem}
\label{th:chebnet}
Chebnet is more powerful than the 1-WL test if the Laplacian maximum eigenvalues of the non-regular graphs to be compared are not the same. Otherwise Chebnet is not more powerful than 1-WL.
\end{theorem}



\begin{figure}[ht]
\begin{center}
\centerline{\includegraphics[width=0.6\columnwidth]{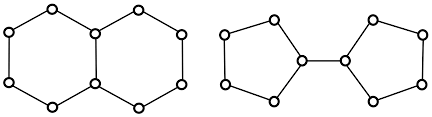}}
\caption{Decalin ($G$) and Bicyclopentyl ($H$) graphs are $\mathcal{L}_1$ and also 1-WL equivalent, but Chebnet can distinguish them.
}
\label{dec_bicyc}
\end{center}
\vskip -0.2in
\end{figure}

Figure~\ref{dec_bicyc} shows two graphs that are 1-WL equivalent and are generally used to show how MPNNs fail. However, their normalized Laplacian's maximum eigenvalues are not the same. Thus, Chebnet can project these two graphs to different points in feature space. Details can be found in Appendix~\ref{app2}.

\begin{figure}[ht]
\begin{center}
\centerline{\includegraphics[width=0.8\columnwidth]{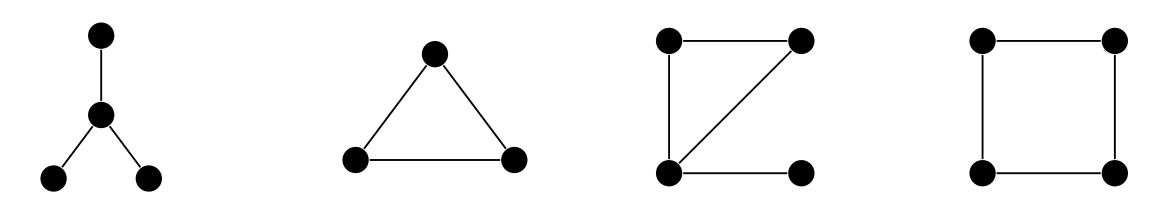}}
\caption{Sample of patterns: 3-star, triangle, tailed triangle and 4-cycle graphlets used in our analysis.
}
\label{substruct}
\end{center}
\vskip -0.2in
\end{figure}

As stated in the introduction, comparison with the WL-test is not the only way to characterize the expressive power of GNNs. Powerful GNNs are also  expected to be able to count relevant substructures in a given graph for specific problems. The following theorems describe the matrix language required to be able to count the graphlets illustrated in Figure~\ref{substruct}, which are called 3-star, triangle, tailed triangle and 4-cycle. 

\begin{theorem}
\label{th:star3}
3-star graphlets can be counted by sentences in  $\mathcal{L}_1^+$.
\end{theorem}

\begin{theorem}
\label{th:triangle}
Triangle and 4-cycle graphlets can be counted by sentences in  $\mathcal{L}_2^+$.
\end{theorem}

\begin{theorem}
\label{th:tailed4cycle}
Tailed triangle graphlets can be counted by sentences in  $\mathcal{L}_3^+$.
\end{theorem}

These theorems show that 1-WL equivalent MPNNs can only count 3-star patterns, while 3-WL equivalent MPNNs can count all graphlets shown in Figure~\ref{substruct}. 

\cite{DehmamyBY19} has shown that a MPNN is not able to learn node degrees if the MPNN has not an appropriate convolution support (e.g. $A$). Therefore, to achieve a fair comparison, we assume that node degrees are included as a node feature. Note however, that the number of 3-star graphlets centered on a node can be directly derived from its degrees (see Appendix~\ref{section:prth3}). Therefore, any graph agnostic MLP can count the number of 3-star graphlets given the node degree. 


\section{MPNN Beyond 1-WL}

In this section, we present two new MPNN models. The first one, called GNNML1 is shown to be as powerful as the 1-WL test. The second one, called GNNML3 exploits the theoretical results of  \cite{geerts2020expressive} to break the limits of 1-WL and reach 3-WL equivalence experimentally. 
GNNML1 relies on the node update schema given by :
\begin{equation}
    \scriptstyle
  \label{eq:ml1}
  H^{(l+1)}=\sigma\left( H^{(l)}W^{(l,1)}+AH^{(l)}W^{(l,2)} \\ + H^{(l)}W^{(l,3)}\odot H^{(l)}W^{(l,4)}  \right)
\end{equation}
where $W^{(l,s)}$ are trainable parameters. Using this model, the new representation of a node consists of a sum of three terms : (i) a linear transformation of the previous layer representation of the node, (ii) a linear transformation of the sum of the previous layer representations of its connected nodes and (iii) the element-wise multiplication of two different linear transformations of the previous layer representation of the node.

The expressive power of GNNML1 is defined by the following theorem. Its proof is given in Appendix \ref{section:app_prof}:
\begin{theorem}
\label{th:th1}
GNNML1 can produce every possible sentences in $ML(\mathcal{L}_1)$ for undirected graph adjacency $A$ with monochromatic edges and nodes. Thus, GNNML1 is exactly as powerful as the 1-WL test.
\end{theorem}

Hence, this model has the same ability as the 1-WL test to distinguish two non-isomorphic graphs, i.e., the same as GIN. This is explained by the third term in the sum of Eq.\eqref{eq:ml1} since it can produce feature-wise multiplication 
on each layer. Since node representation is richer, we also assume that it would be more powerful for counting substructures. This assumption is validated by experiments in Section~\ref{section:result}.

To reach more powerful models than 1-WL, theoretical results 
(see Remarks~\ref{rm:rm2}, \ref{rm:rm3} and \ref{rm:rm5} in Section \ref{section:WL}) show that a model that can produce different outputs than $\mathcal{L}_1^+$ language is needed. More precisely, according to Remarks~\ref{rm:rm3} and \ref{rm:rm5}, trace ($tr$) and element-wise multiplication ($\odot$) operations are required to go further than 1-WL. 

In order to illustrate the impact of the trace operation, one can use 1-WL equivalent Decalin and Bicyclopentyl graphs in Figure~\ref{dec_bicyc}. It is easy to show that $tr(A_G^5)=0$ but $tr(A_H^5)=20$, $tr(A^5)$ giving the number of 5-length closed walks. Thus, if a model can apply a trace operator over some power of adjacency, it can easily distinguish these two graphs. Computational details concerning this example are given in Appendix~\ref{app2}. 

\begin{figure}[ht]
\begin{center}
\centerline{\includegraphics[width=0.5\columnwidth]{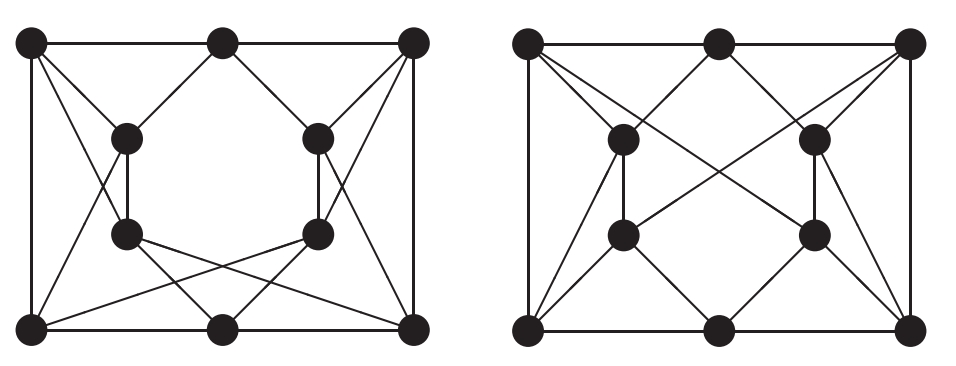}}
\caption{Cospectral and 4-regular graphs from \cite{van2003graphs} are $\mathcal{L}_1$ and $\mathcal{L}_2$ equivalent.
}
\label{cospect}
\end{center}
\vskip -0.4in
\end{figure}

Despite this interesting property of the trace operator, it is not sufficient to distinguish cospectral graphs, since cospectral graphs (see  Figure~\ref{cospect}) have the same number of closed walks of any length (see Proposition 5.1 in \cite{geerts2020expressive}). In such cases, element-wise multiplication is useful. As an example, the sentence $e(A)={\bf1}^{\top}f((A\odot A^2)^2{\bf1})$  where $f(x)=x \odot x$ for any vector $x$, gives $e(A_G)=6032$ and $e(A_H)=5872$ for the graphs of Figure~\ref{cospect}. Thus, element-wise multiplication helps distinguishing these two graphs. The calculation details can be found in Appendix~\ref{app3}.

As shown by these examples, a model enriched by element-wise multiplication and trace operator can go further than the 1-WL test. However, these operations need to keep the power of the adjacency matrix explicitly and to multiply these dense matrices to each other by matrix or element-wise multiplication. Such a strategy is actually used by higher order GNNs such as \cite{maron2019provably, 123lehmango}, which are provably more powerful than existing MPNNs. 

However, MPNNs cannot calculate the power of a given adjacency explicitly. Indeed, a MPNN layer multiplies the previous representation of the nodes by sparse adjacency matrix or more generally sparse convolution supports $C$ in Eq.\eqref{eq:mpgnn}. More precisely, if the given node features are $H^{(0)}={\bf1}$, a MPNN can calculate $C^3{\bf1}$ by 3 layered MPNN computing $C(C(C{\bf1}))$ but not by $(C^3){\bf1}$. Since a MPNN does not keep $C^3$ explicitly, it cannot take its trace or multiply element-wise to another power of support. This is a major disadvantage of MPNNs, but it explains why MPNNs need just linear time and memory complexity, making them useful in practice.

A solution to the problem mentioned above is to design graph convolution supports by the element-wise multiplication of the $s$-power of the adjacency matrix and a given receptive field, i.e., by $C^{(s)}=M\odot A^s$ where $M$ masks the components of the powered matrix and keeps the convolution support sparse. $M=A+I$ is an example of mask that gives a maximum 1-length receptive field. This model cannot calculate all possible element-wise multiplications between all possible matrices, but it can produce any sentence in a form of $(M\odot A^s)^l$ where $l \in [0,l_{max}]$ is the layer number and $s \in [0,s_{max}]$ is the pre-computed power of convolution supports. In this proposition, the receptive field mask and the number of power of adjacency should be computed in a pre-processing step. However, we cannot initially know  which power of adjacency matrix is necessary for a given problem. One solution is to tune it as an hyperparameter of the model. Another problem of this approach is that using powers of adjacency makes the convolution supports filled with high values that have to be normalized.

To overcome these problems, we propose through our GNNML3 model to design convolution supports in the spectral domain as functions of eigenvalues of the normalized Laplacian matrix or of the adjacency matrix. The following theorem, with proof given in Appendix \ref{section:app_prof}, shows that such supports can be written as power series of the graph Laplacian or the adjacency matrix. 

\begin{theorem}
\label{th:th4}
A convolution support given by
\begin{equation}
  \label{eq:th4eq11}
  C'^{(s)}=U diag(\Phi_s(\lambda))U^{\top},
\end{equation}
where $\Phi_s(\lambda)=exp(-b(\lambda-f_s)^2)$, $f_s\in [\lambda_{min},\lambda_{max}]$ is a scalar design parameter of each convolution support and $b >0$ is a general scalar design parameter, 
can be expressed as a linear combination of all powers of graph Laplacian (or adjacency) as follows, with $\alpha_{s,i}=\frac{\Phi_s^{(i)}(0)}{i!}$: 
\begin{equation}
  \label{eq:th4eq12}
  C'^{(s)}=\alpha_{s,0}L^0 +\alpha_{s,1}L^1 +\alpha_{s,2}L^2 +  \dots.
\end{equation}
\end{theorem}
Since design parameters $f_s$ of each matrix are different, each $C'^{(s)}$ in Eq.\eqref{eq:th4eq12} consists of different linear combinations of power series of the graph Laplacian (or adjacency). Thus, necessary powers of the graph Laplacian (or adjacency) and its diagonal part (for trace operation) can be learned and their element-wise multiplication can be produced by: 
\begin{equation}
  \label{eq:learnsup}
  C=M \odot mlp_{4}\left (mlp_{1}(C') | mlp_{2}(C') \odot mlp_{3}(C')\right ),
\end{equation}
where $C'=[C'^{(1)}|\dots|C'^{(s)}] \in \mathbb{R}^{n\times n \times S}$ stacks initial convolution supports on the third dimension, $mlp_{k}(.)$ is a trainable model performed on third dimension of a given tensor, $C=[C^{(1)}|\dots|C^{(S)}] \in \mathbb{R}^{n\times n \times S}$ sparsify convolution support by defined receptive field mask $M$. The forward calculation of one layer MPNN becomes:
\begin{equation}
  \label{eq:ml2}
  \footnotesize
  H^{(l+1)}=\sigma\Big( \sum_s(C^{(s)}H^{(l)}W^{(l,s)}) | mlp_5(H^{(l)})\odot mlp_6(H^{(l)})  \Big)
\end{equation}   
where we concatenate MPNN representation under learned convolution with element-wise product of node representations as in GNNML1. 

There is an infinite number of selections of $\Phi_s(\lambda)$ that make the convolution support written by power series of graph Laplacian (or adjacency). 
However, we can design each convolution support to be sensitive on each band of spectrum ($f_s$) by given bandwidth ($b$). Therefore, our model will be able to learn properties depending on the spectrum of graph signal.


Algorithm~\ref{alg:preprocapplicationes} 
calculates the initial convolution supports ($C'$). Since the supports are computed for valid indices in the receptive field mask (where $M_{i,j}=1$), one can see $C'$ as extracted edge features where the edge indices are defined by $M$. In application, a function $sparse2vec: \mathbb{R}^{n \times n} \rightarrow \mathbb{R}^{m}$ converts the sparse matrix to a vector by just keeping the components on valid indices of the mask. 
\begin{algorithm}[tb]
  \caption{GNNML3 Preprocessing Step}
  \label{alg:preprocapplicationes}
  \footnotesize
\begin{algorithmic}
  \STATE {\bfseries Input:} adjacency $A \in \mathbb{R}^{n \times n}$, receptive field mask $M \in \{0,1\}^{n \times n}$, number of supports $S \in \mathbb{N}$, frequency responses function of each support $\Phi_1(\lambda) \dots \Phi_S(\lambda)$
  \STATE {\bfseries Output:} extracted edge features $C' \in \mathbb{R}^{m \times S}$
  \STATE Set basis matrix: $B=I-D^{-1/2}AD^{-1/2}$ or $B=A$.
  \STATE Eigendecomposition: $Udiag(\lambda)U^{\top}=B$
  \FOR{$s=1$ {\bfseries to} $S$}
  \STATE $C'_{:s}=sparse2vec\left( M \odot (U diag(\Phi_s(\lambda))U^{\top})\right)$
  \ENDFOR
\end{algorithmic}
\end{algorithm}
Algorithm~\ref{alg:proposed} shows the forward calculation of the model for just one graph. To make the representation as simple as possible, we prefer to use tensor representation in Eq.\eqref{eq:learnsup}.
However, implementation of Algorithm~\ref{alg:proposed} just apply $mlp_{k}(.)$ to the valid indices defined by receptive field mask $M$. Thus, $C,C'$ have the dimension of $\mathbb{R}^{m \times S}$, where $m$ shows number of valid indices in $M$ and $mlp_{k}(.)$ applies on columns of $C'$. Beside, we use a function $vec2sparse: \mathbb{R}^{m} \rightarrow \mathbb{R}^{n \times n}$ that converts the vector to the sparse convolution support according to a given mask $M$.

The limit of the proposed method is similar to the limit of 3-WL (or 2-FWL) test. For instance, it fails to distinguish strongly regular graphs, that can be defined by 3 parameters: the degree of the nodes, the number of common neighbours of adjacent node pairs,  and the number of common neighbours of non-adjacent node pairs. Such graphs are provably known to be 3-WL equivalent \cite{arvind2020weisfeiler}. In Appendix~\ref{app4}, a strongly regular graphs pair and the result of a sample sentence in $\mathcal{L}_3$ are presented.

\begin{algorithm}[tb]
\caption{GNNML3 Forward calculation}
  \label{alg:proposed}
  \footnotesize
\begin{algorithmic}
  \STATE {\bfseries Input:} extracted edge features
  $C' \in \mathbb{R}^{m  \times S}$, initial node feature $H^{(0)} \in \mathbb{R}^{n \times f_0}$, receptive field mask $M \in \{0,1\}^{n \times n}$, number of layers $L$, number of supports $S$
  \STATE {\bfseries Output:} new node representation $H^{(L)}$
  \FOR{$l=0$ {\bfseries to} $L-1$}
  \STATE $\scriptstyle \Tilde{C}=mlp_{l,4}(mlp_{l,1}(C') | mlp_{l,2}(C') \odot mlp_{l,3}(C'))$
  \FOR{$s=1$ {\bfseries to} $S$}
  \STATE
  $\scriptstyle C^{(s)}=vec2sparse(\Tilde{C}_{:s},M)$
  \ENDFOR
  \STATE
  $\scriptstyle
  H^{(l+1)}=\sigma\left( \sum_s(C^{(s)}H^{(l)}W^{(l,s)}) | mlp_{l,5}(H^{(l)})\odot mlp_{l,6}(H^{(l)}) \right)$
  \ENDFOR
\end{algorithmic}
\end{algorithm}
\section{Experimental Results}
\label{section:result}

This section presents the experimental results obtained by the proposed models GNNML1 and GNNML3. All codes and datasets are available online \footnote{https://github.com/balcilar/gnn-matlang}. We use GCN, GAT, GIN and Chebnet as 1-WL MPNN baselines and PPGN as 3-WL baseline (see Appendix~\ref{section:implementation}). Experiments aim to answer four questions:

\textbf{Q1}: How many pairs of non-isomorphic simple graphs that are either 1-WL or 3-WL equivalent are not distinguished by the models? \\
    \textbf{Q2}: Can the models generalize the counting of some substructures in a given graph? \\
    \textbf{Q3}: Can the models learn low-pass, high-pass and band-pass filtering effects and  generalize the classification problem according to the frequency of the signal? \\
    \textbf{Q4}: Can the models generalize downstream graph classification and regression tasks?


In order to perform experimental expressive power tests, we use graph8c and sr25 datasets\footnote{http://users.cecs.anu.edu.au/$\sim$bdm/data/graphs.html}. Graph8c is composed of all the \num{11117} possible connected non-isomorphic simple graphs with 8 nodes. We compare all possible pairs of graphs of this dataset, leading to more than 61M comparisons. According to our test, we found that 312 pairs out of 61M are 1-WL equivalent and none of the pairs are 3-WL equivalent.
The sr25 dataset contains strongly regular graphs where each graph has 25 nodes, each node's degree is 12, connected nodes share 5 common neighbours and non-connected nodes share 6 common neighbors. Sr25 consists of 15 graphs, leading to 105 different pairs for comparison.

Moreover, we use the EXP dataset~\cite{abboud2020surprising}, having 600 pairs of 1-WL equivalent graphs. This dataset also includes a binary classification task. Depending on graph features, each graph of a pair of 1-WL equivalent graphs is assigned to two different classes. We split the dataset
into 400, 100, and 100 pairs for train, validation and test sets respectively. The test set is used to measure the generalization ability: a model that fails to distinguish 1-WL equivalent graphs inevitably fails to learn this task.

We use 3-layer graph convolution followed by sum readout layer, and then a linear layer to convert the readout layer representation into a 10-length feature vector. We keep the parameter budget around 30K for all methods. For graph8c, sr25 and EXP tasks, there is no learning. Model weights are randomly initialized and 10-length graph representations are compared by the Manhattan distance. If the distance is less than $10^{-3}$ in all 100 independent runs
, we assume the pairs are similar. For EXP-classification task, we train the model and pick the best one according to validation set performance and report its performance on test set.

\begin{table}[t]
\caption{Number of undistinguished pairs of graphs in graph8c, sr25 and EXP datasets and binary classification accuracy on EXP dataset. An ideal method does not find any pair similar and classifies graphs with 100\% accuracy. The number of pairs is 61M for graph8c, 105 pairs for sr25 and 600 for EXP.}
\label{exp8c}
\vskip 0.15in
\begin{center}
\begin{scriptsize}
\begin{sc}
\begin{tabular}{lcccc}
\toprule
Model & graph8c & sr25 & EXP & EXP-classify \\
\midrule
MLP    & 293K & 105 & 600 & 50\% \\
GCN & 4775 & 105 & 600 &  50\%\\
GAT    & 1828 & 105 &600 & 50\%\\
GIN    & 386 & 105 & 600 &50\%\\
Chebnet     & 44 & 105 & 71 &82\%\\
PPGN      & \textbf{0}& 105 & \textbf{0} & \textbf{100\%} \\
\textbf{GNNML1}      & 333 & 105 & 600 & 50\%\\
\textbf{GNNML3}   & \textbf{0}& 105 & \textbf{0} & \textbf{100\%}\\
\bottomrule
\end{tabular}
\end{sc}
\end{scriptsize}
\end{center}
\vskip -0.2in
\end{table}

Table~\ref{exp8c} presents the obtained results. One can see that 99.5\% of the graphs in graph8c dataset can be distinguished even by graph agnostic method MLP (293K out of 61M is not separable by MLP). This can be explained by the fact that the node degrees has been added as node features. Hence, all methods initially know the result of first iteration of 1-WL test. Thus, MLP (and also first iteration of 1-WL test) can distinguish pairs of graphs when multiset of node degrees are not same. GNNML1 and GIN's result is very closed to the theoretical limit of 1-WL test which is 312 pairs for graph8c dataset. The difference can be explained by threshold value to make decision if the two representations are equal and/or the number of layers in the model. It is possible that 1-WL test may need more than 3 iteration to distinguish some pairs. 
Due to having less expressive power of GCN and GAT compare to the 1-WL test, their performances are worse than 1-WL test. Since graph8c dataset has 1-WL equivalent non-regular graph pairs that have different maximum eigenvalue, Chebnet could detect these pairs and reaches better performance than theoretical limit of 1-WL test as stated by Theorem~\ref{th:chebnet}.  

On EXP dataset, composed of 1-WL equivalent graph pairs, MPNNs cannot distinguish any pair of graphs, except Chebnet which is able to distinguish all the pairs with different maximum eigenvalues. In EXP there is no regular graphs and only 71 graph pairs have similar maximum eigenvalues. Chebnet fails on these pairs but distinguishes the others, as stated by Theorem~\ref{th:chebnet}. 
One can note that using a fixed value for maximum eigenvalue (e.g. $\lambda_{max}=2$ as it is usually done in practice) reduces Chebnet performance to those of MPNNs. 

Similarly to results on EXP, 1-WL equivalent MPNNs except Chebnet fail to predict of EXP classification task and do not perform better than random prediction. On the contrary, PPGN and GNNML3 have perfect results on graph8c, EXP and EXP-classify tasks thanks to their 3-WL equivalence. However, since strongly regular graphs are 3-WL equivalent, no model less or as powerful as 3-WL test can distinguish the pairs in sr25 dataset. To obtain a better result on this dataset, we need to go further than 3-WL (see Appendix~\ref{app4}). These experiments reply to \textbf{Q1}.

To bring an answer to \textbf{Q2}, we propose to count 3-star, triangle, tailed-triangle and 4-cycle substructures (Fig.~\ref{substruct}). In addition to these 4 graphlets, we also create another task (noted as \textsc{CUSTOM} in Table~\ref{graphletcount}) that aims to approximate a custom sentence $e_c \in \mathcal{L}_1^+$ ,  $e_c(A) = {\bf1}^{\top} A\,diag(exp(-A^2{\bf1}))A{\bf1}$ with $A$ the graph adjacency matrix. Since $e_c \in ML(\mathcal{L}_1^+)$, it may be learnable by 1-WL equivalent MPNNs. We used the RandomGraph dataset~\cite{chen2020can} with same partitioning: 1500, 1000 and 2500 graphs for train, validation and test respectively. To create the ground truth of number of graphlets, we count them  according to theorem proofs in Appendix~\ref{section:prth3}, \ref{section:prth4}, \ref{section:prth5} and normalized the number to a unitary standard deviations, to keep the errors in the same scale as in Table~\ref{graphletcount}. We use 4 convolution layers, a graph readout layer computing a sum and followed by 2 fully connected layers. All methods parameter budget is around 30K. We keep the maximum number of iterations to  200 and we stop the algorithm if the error goes below \num{e-4}. 

\begin{table}[t]
\caption{Median of test set MSE error for graphlet counting problem on random graph dataset over 10 random runs.}
\label{graphletcount}
\vskip 0.15in
\begin{center}
\begin{scriptsize}
\begin{sc}
\begin{tabular}{lccccc}
\toprule
Model & 3-stars &custom & triangle & tailed-tri & 4-cycles \\
\midrule
MLP    & \textbf{1.0E-4} &4.58E-1  & 3.13E-1 & 2.22E-1 & 1.73E-1 \\
GCN & \textbf{1.0E-4} &3.22E-3& 2.43E-1 & 1.42E-1 &  1.14E-1\\
GAT    & \textbf{1.0E-4} & 4.57E-3&2.47E-1 &1.44E-1 & 1.12E-1\\
GIN    & \textbf{1.0E-4} & 1.47E-3&2.06E-1 & 1.18E-1 &1.21E-1\\
Chebnet     & \textbf{1.0E-4} &\textbf{7.68E-4}& 2.01E-1 & 1.15E-1 &9.60E-2\\
PPGN      & \textbf{1.0E-4} & \textbf{9.19E-4}& \textbf{1.00E-4} & \textbf{2.61E-4} & \textbf{3.30E-4} \\
\textbf{GNNML1}      & \textbf{1.0E-4} & \textbf{2.75E-4}& 2.45E-1 & 1.32E-1 & 1.14E-1\\
\textbf{GNNML3}   & \textbf{1.0E-4} & \textbf{7.24E-4}& \textbf{4.44E-4} & \textbf{3.18E-4} & \textbf{6.62E-4} \\
\bottomrule
\end{tabular}
\end{sc}
\end{scriptsize}
\end{center}
\vskip -0.1in
\end{table}

The results in Table~\ref{graphletcount} are consistent with Theorems~\ref{th:star3},  \ref{th:triangle}, \ref{th:tailed4cycle}. 3-WL models are able to count graphlets and approximate our custom function (result $<$ \num{E-3}), while 1-WL equivalent models can only count the 3-stars graphlet, as stated in Theorem~\ref{th:star3}. Custom function approximation results also show that GNNML1 and Chebnet provide better approximation of the target other MPNNs, which is again consistent with our analysis.

Question \textbf{Q3} concerns the spectral expressive power of models. Such an analysis is important when input-output relations depend on the spectral properties of the graph signal such as in image/signal processing applications. As shown in \cite{balcilar2021analyzing}, the vast majority of existing MPNNs operate as low-pass filters which limits their capacity. To lead this analysis, we use the datasets presented in \cite{balcilar2021analyzing}. First, we evaluate if the models can learn low-pass, high-pass and band-pass filtering effects, through a node regression problem. Model performances are thus reported $R^2$ using mean square error (MSE) loss. The original data consists in a 2-d grid graph of size 100x100. Since the PPGN's memory and computational complexity is prohibitive with a reasonable computer, 
we select 3 different 30x30 regions of the original 2-d grid graph  as training, validation and test sets. 
A second dataset consists of 5K planar graphs, split into 3K, 1K and 1K sets for train, validation and test. They are used to evaluate if the models can classify graphs into binary classes where the ground truth labels were determined according to the frequency of the signal on the graph. Since the problem is binary graph classification we use binary cross entropy loss. 

\begin{table}[t]
\caption{Spectral expressive analysis results. $R^2$ for LowPass, HighPass and BandPass node regression tasks, accuracy on graph classification task. Results are median of 10 different runs.}
\label{spect}
\vskip 0.15in
\begin{center}
\begin{scriptsize}
\begin{sc}
\begin{tabular}{lcccc}
\toprule
Model & Low-Pass & High-Pass & Band-Pass & Classify \\
\midrule
MLP    & 0.9749 & 0.0167  & 0.0027 & 50.0\% \\
GCN  & 0.9858 & 0.0863 & 0.0051 & 77.9\% \\
GAT   & 0.9811 & 0.0879 & 0.0044 & 85.3\% \\
GIN   & 0.9824 & 0.2934 & 0.0629 & 87.6\% \\     
Chebnet     & \textbf{0.9995} & \textbf{0.9901} & \textbf{0.8217} & \textbf{98.2\%} \\
PPGN      & \textbf{0.9991} & \textbf{0.9925} & 0.1041 & 91.2\% \\
\textbf{GNNML1}   & \textbf{0.9994} & \textbf{0.9833} & 0.3802 & 92.8\% \\
\textbf{GNNML3}  & \textbf{0.9995} & \textbf{0.9909} & \textbf{0.8189} & \textbf{97.8\%} \\
\bottomrule
\end{tabular}
\end{sc}
\end{scriptsize}
\end{center}
\vskip -0.1in
\end{table}

The results of spectral expressive power analysis are presented in Table~\ref{spect}. Node regression results show that 1-WL equivalent existing MPNNs can mostly learn low-pass effects. By applying different weights to self node and neighbourhood, GNNML1 can learn high pass effect relatively well. PPGN also learns high-pass effect  better than 1-WL equivalent methods. Band-pass can be generalized by Chebnet and GNNML3 thanks to the convolutions designed in spectral domain. The reason why the band-pass regression results are worse than the low and high-pass results is that the ground truth band-pass effect is created by very stiff frequency function and Chebnet also GNNML3 need more convolution supports to learn it. Because of non-local process in PPGN, it cannot learn the band-pass effect and provide no better result than 1-WL MPNNs in graph classification problem. Thus, Chebnet and GNNML3 give the best results on all spectral ability test, thanks to their spectral convolutions process. 

For answering the last question \textbf{Q4}, we apply the different models on some common benchmark tasks and datasets. Table~\ref{task} and Table~\ref{TUtask} present the performance of both baseline models and the proposed ones on these benchmark datasets. The results on Zinc12K and \textsc{MNIST}-75 datasets are very interesting because of the nature of these two problems. The solution of the Zinc12K dataset mostly depends on structural features of the graph. For instance, a recent study reaches 0.14 MAE  by using handcrafted features, 
which cannot be extracted by a 3-WL equivalent model~\cite{bouritsas2020improving}. Obtained results confirm that models that are able to count substructures, such as PPGN and GNNML3, perform better than others with a large margin. On the other hand, since \textsc{MNIST}-75 dataset is based on image analysis, it needs a model with a higher spectral ability. Therefore, Chebnet and GNNML3 perform significantly better than other models on this task. Our proposal GNNML3 gives comparable results on other TU datasets in \cite{morris2020tudataset} such as MUTAG, ENZYMES, PROTEINS and PTC presented in Appendix~\ref{section:TUresult}. 

\begin{table}[t]
\caption{Results on Zinc12K and MNIST-75 datasets. The values are the MSE for Zinc12K and the accuracy for MNIST-75. Edge features are not used even if they are available in the datasets. For Zinc12K, all models use node labels. For MNIST-75, the model uses superpixel intensive values and node degree as node features.}
\label{task}
\vskip 0.15in
\begin{center}
\begin{scriptsize}
\begin{sc}
\begin{tabular}{lcc}
\toprule
Model & Zinc12K & MNIST-75   \\
\midrule
MLP     & 0.5869 $\pm$ 0.025& 25.10\% $\pm$ 0.12  \\
GCN     & 0.3322 $\pm$ 0.010 & 52.80\% $\pm$ 0.31  \\
GAT     & 0.3977 $\pm$ 0.007 & 82.73\% $\pm$ 0.21  \\
GIN     & 0.3044 $\pm$ 0.010 & 75.23\% $\pm$ 0.41  \\  
Chebnet & 0.3569 $\pm$ 0.012 & \textbf{92.08\% $\pm$ 0.22}  \\
PPGN    & \textbf{0.1589 $\pm$ 0.007} & 90.04\% $\pm$ 0.54  \\
\textbf{GNNML1}  & 0.3140 $\pm$ 0.015 & 84.21\% $\pm$ 1.75 \\
\textbf{GNNML3}  & \textbf{0.1612 $\pm$ 0.006} & \textbf{91.98\% $\pm$ 0.18}  \\
\bottomrule
\end{tabular}
\end{sc}
\end{scriptsize}
\end{center}
\vskip -0.1in
\end{table}

\section{Conclusion}


Despite a computational and memory efficiency, MPNN is known to have an expressive power limited to 1-WL test. MPNN is then unable to distinguish 1-WL equivalent graphs and cannot count some substructures of the graph. In this paper, we have presented new models, by translating the insights of MATLANG to the GNN world. This solution gives access to a new MPNN that is theoretically more powerful than the 1-WL test, and experimentally as powerful as 3-WL existing models for distinguishing non-isomorphic graphs and for counting substructures without feature engineering nor node permutations in the training phase. The proposed MPNN is also powerful in terms of spectral expressive ability, going beyond low-pass filtering, which is another expressive perspective of GNNs. Experimental results confirm the theorems stated in the paper. The proposed method has a big advantage over all studied MPNN on graph isomorphism and substructure counting tasks. With respect to the 3-WL equivalent baseline PPGN, the biggest advantage of our proposal is its complexity. Proposed GNNML3 needs linear memory and time complexity with respect to the number of nodes, while PPGN needs quadratic memory and cubic time complexity, making the model infeasible for large graphs. The second advantage over PPGN is that since it is created in the spectral domain, its convolution process takes care of signal frequencies, making it more efficient in terms of output signal frequency profile. 

\section*{Acknowledgments}

This work was partially supported by the Normandy Region (grant RiderNet), the French Agence National de Recherche (grant APi, ANR-18-CE23-0014) and the PAUSE Program of Coll\`ege de France.

\bibliography{main}
\bibliographystyle{icml2021}

\clearpage

\appendix
\section{Weisfeiler-Lehman Test}
\label{section:app_wl}

The universality of a GNN is based on its ability to embed two non-isomorphic graphs to distinct points in the target feature space. A model which can distinguish all pairs of non-isomorphic graphs is a universal approximator. Since it is not known if the graph isomorphism problem can be solved in polynomial time or not, this problem is neither NP-complete nor P, but NP-intermediate \cite{takapoui2016linear}. One of the oldest but prominent polynomial approach is the Weisfeiler-Lehman Test (abbreviated WL-test) which gives 
sufficient but not enough evidence. WL test can be extended by taking into account higher order of node tuple within the iterative process. These extensions are denoted as $k$-WL test, where $k$ is equal to the order of the tuple. It is important to mention that an higher order of tuple leads to a better ability to distinguish two non-isomorphic graphs (with the exception for $k=2$) ~\cite{arvind2020weisfeiler}.


The 1-WL test, known as vertex coloring, starts with the given initial color of nodes if available. Otherwise all nodes are colored with the same color ($H_v^{(0)}=1$). Then, colors are updated by the following iteration:
\begin{equation}
  \label{eq:wl1}
  H_v^{(t+1)} =\sigma \left( H_v^{(t)}~ |\left\{ H_u^{(t)} : u \in \mathcal{N}(v) \right\} \right),
\end{equation}
where $H_v^{(t)}$ is the color of vertex $v$ at iteration $t$, $\mathcal{N}(v)$ is the set of neighbours of vertex $v$, $|$  represents the concatenation operator and $\{.\}$ is the order invariant multiset\footnote{It is 
generally implemented 
by stacking all colors in the set and sorting them alphabetically}. In order to avoid the new color of vertex become bigger after each iteration due to the concatenation operation and to keep the color description simple, the recoloring $\sigma(\cdot)$ function is applied after each iteration. It assigns a new simple color identifier to the any newly created color. The test is performed in parallel for two graphs. The iterative process is stopped when the color histograms are kept unchanged between two consecutive iterations.
The color histograms associated to the compared graphs are examined. If in any iteration the histograms are different, we can conclude that the graphs are not isomorphic. However, the opposite conclusion can not be drawn if color histograms are equal as two same histograms may be computed even for non-isomorphic graphs.

Higher order WL tests use the same algorithm while their color update schema is slightly different. The 2-WL test uses second order tuple of nodes (all ordered pairs of nodes), thus it needs $\mathbf{H} \in \mathbb{R}^{n \times n}$ matrix, where the initial color set has two more colors than initial vertex colors as defined by:
\begin{equation}
  \label{wl2initial}
  \mathbf{H}_{v,u}^{(0)} = \left\{
        \begin{array}{l@{\text{ ~~~~if~ }}l}
            H_v^{(0)} & v = u \\           
            edge & u \in \mathcal{N}(v) \\  
            nonedge & u \not\in \mathcal{N}(v) \\
        \end{array}
    \right. 
\end{equation}
Then, the iteration process is applied through the following schema where $[n]$ is the set of node identifiers.
\begin{equation}
  \label{eq:wl2}
  \mathbf{H}_{v,u}^{(t+1)} =\sigma \left( \mathbf{H}_{v,u}^{(t)}~ |\left\{ \mathbf{H}_{v,k}^{(t)} : k \in [n] \right\} |\left\{ \mathbf{H}_{k,u}^{(t)} : k \in [n] \right\} \right),
\end{equation}
Although for $k\geq2$, $(k+1)$-WL is more powerful than $(k)$-WL, it is not true for $k=1$, thus 2-WL (Eq.\eqref{eq:wl2}) is no more powerful than 1-WL  (Eq.\eqref{eq:wl1}) \cite{maron2019provably}. To clarify this point,
Folkore WL (FWL) test is defined such that 1-WL=1-FWL, but for $k \geq 2$, we have $(k+1)$-WL $\approx$ $(k)$-FWL~\cite{maron2019provably}.
The iteration process of 2-FWL is given by the following equation;
\begin{equation}
  \label{eq:fwl2}
  \mathbf{H}_{v,u}^{(t+1)} =\sigma \left( \mathbf{H}_{v,u}^{(t)}~ |\left\{ \left( \mathbf{H}_{v,k}^{(t)} | \mathbf{H}_{k,u}^{(t)} \right) : k \in [n]  \right\} \right),
\end{equation}

In the literature, there are different interpretations of the order of the WL test. Some papers use WL test order to denote the iteration given by Eq.\eqref{eq:wl1} and Eq.\eqref{eq:wl2} \cite{123lehmango, maron2019provably} but some others such as \cite{abboud2020surprising,arvind2020weisfeiler,takapoui2016linear} use FWL order under the name of WL. In this paper, we explicitly mention both WL and FWL equivalent such as 3-WL (or 2-FWL) to alleviate ambiguities.

\section{Proofs of Theorems}
\label{section:app_prof}

\subsection{Theorem.1}

\begin{proof}
\label{section:prth1}

All these methods can be written in Eq.\eqref{eq:mpgnn} by different convolution matrices $C$. The main idea of the proof is that as long as convolution matrices $C$ can be explained by operations from the enriched set $\mathcal{L}_1^+$ (Remark~\ref{rm:rm4}), Eq.\eqref{eq:mpgnn} also can be explained by operations from $\mathcal{L}_1^+$ as well. Thus these methods cannot produce any sentence out of $\mathcal{L}_1^+$. As a consequence, their expressive power is not more than 1-WL test. To provide a proof, the mentioned methods' convolution matrices have to be expressed using operations from $\mathcal{L}_1^+$.

GCN uses $C=(D+I)^{-0.5}(A+I)(D+I)^{-0.5}$ where $D$ is the diagonal degree matrix \cite{kipf16:_semi} in Eq.\eqref{eq:mpgnn}. $(D+I)^{-0.5}$ can be expressed as $(D+I)^{-0.5}=diag(f(A{\bf1}+{\bf1}))$, where $f(x)=x^{-0.5}$ is element-wise operation on vector $x$.  $A+I$ can also be written $A+diag({\bf1})$. When we merge these equations, we get $C=diag(f(A{\bf1}+{\bf1}))(A+diag({\bf1}))diag(f(A{\bf1}+{\bf1}))$. The convolution support $C$ is then written using operations from $\mathcal{L}_1^+$.  

In the literature, GraphSage method was proposed to sample neighborhood and aggregate the neighborhood contribution by the mean operator or LSTM in \cite{hamilton2017sage}. Since we restrict the method using full sampling and mean aggregator, we can define GraphSage by the general framework given by Eq.\eqref{eq:mpgnn} with two convolution supports which are the identity matrix $C^{(1)}=I$ and the row normalized adjacency matrix $C^{(2)}=D^{-1}A$. These convolution supports can also be expressed by operations from $\mathcal{L}_1^+$, by observing that $C^{(1)}=diag({\bf1})$ and $C^{(2)}=diag(f(A{\bf1}))A$, where $f(x)=x^{-1}$ elementwise operation on vector $x$.

GIN \cite{xu2018how} uses a convolution support $C=A+I\epsilon$ in Eq.\eqref{eq:mpgnn} which is followed by a custom number of MLP layers. Each of these layers correspond to a convolution support that can by expressed as $C_{mlp}=I$ in Eq.\eqref{eq:mpgnn}. Finally, these convolution supports can be written thanks to operations from $\mathcal{L}_1^+$. $C=A+\epsilon \times diag({\bf1})$ and $C_{mlp}=diag({\bf1})$.  

GAT \cite{velivckovic2017graph} can be expressed in Eq.\eqref{eq:mpgnn} by the convolution support designed by 
$C_{v,u}=m(H_v,H_u)/\sum_{k \in \Tilde{\mathcal{N}}(v)}m(H_v,H_k)$, where $\Tilde{\mathcal{N}}(v)$ is the self-connection added neighborhood of $v$ and $m(.)$ is any trainable model. If we write the trainable model $m(.)$ as a sum of each node such as $m(H_v,H_u)=f_1(H_v)+f_2(H_u)$, we can define an intermediate matrix $B=diag(f_1(H))(A+I)+(A+I)diag(f_2(H))$. Finally the GAT convolution support can be written by $C=diag((B{\bf1})^{-1})B$ using all operations included within the operation set $\mathcal{L}_1^+$. 
\end{proof}

\subsection{Theorem.2}
\label{section:prth2}

\begin{proof}
Chebnet \cite{defferrard16:_convol_neural_networ_graph_fast} uses desired number $k$ of convolution supports in Eq.\eqref{eq:mpgnn}. As long as these convolutions can be written by operations in $\mathcal{L}_1^+$, we can conclude that Chebnet is no more powerful than 1-WL test. But if at least one convolution cannot be explained in $\mathcal{L}_1^+$, we can say it is more powerful than 1-WL test.

Chebnet's convolution supports are $C^{(1)}=I,~~C^{(2)}=2L/\lambda_{\max}-I,~~C^{(k)}=2C^{(2)}C^{(k-1)} - C^{(k-2)}$. The first support can always be written thanks to an operation from $\mathcal{L}_1$ since $C^{(1)}=diag({\bf1})$. Both normalized and combinatorial graph Laplacian can also be written as $L=diag(A{\bf1})-A$ or $L=diag({\bf1})-diag(f(A{\bf1}))A diag(f(A{\bf1}))$ where $f(x)=x^{-1/2}$ elementwise operation on vector $x$. If $\lambda_{max}$ for both graphs are the same, we can use a constant $\alpha=2/\lambda_{max}$. The second convolution support can then be written as $C^{(2)}=\alpha \times L - diag({\bf1})$. It is then expressed by means of operations from $\mathcal{L}_1^+$. Other convolution supports $C^{(k)}=2C^{(2)}C^{(k-1)} - C^{(k-2)}$ are created by matrix multiplication and subtraction of previous supports which can all be expressed by mean of operations from $\mathcal{L}_1^+$. Thus, if the maximum eigenvalues of tested graphs Laplacians are the same, Chebnet is not more powerful than 1-WL.

However, if the maximum eigenvalues are not the same, $C^{(2)}$ cannot be expressed with the help of the constant value $\alpha$. It means that different coefficients should be used for each graph. For two tested graphs $G$ and $H$, we can write second kernel of Chebnet as $C^{(2)}_G=\alpha_G \times L_G - diag({\bf1})$ and $C^{(2)}_H=\alpha_H \times L_H - diag({\bf1})$. If these two graphs are 1-WL equivalent, any sentence build on $\mathcal{L}_1^+$ applied on these graph is equivalent as well. For instance, we can use the sentences of $e(X)={\bf1}^{\top}X{\bf1}$ with operation in $\mathcal{L}_1^+$. The output of the sentence should be same such $e(L_G)=e(L_H)$ yields ${\bf1}^{\top}L_G{\bf1}={\bf1}^{\top}L_H{\bf1}$. If we assume that Chebnet cannot separate these two graphs, we can calculate one layer ChebNet's output by second support with the same sentence and they should be the same such $e(C^{(2)}_G)=e(C^{(2)}_H)$ yields $\alpha_G{\bf1}^{\top}L_G{\bf1}=\alpha_H{\bf1}^{\top}L_H{\bf1}$. Last equation has contradiction to the previous one as long as the maximum eigenvalues are not same (i.e $\alpha_G \neq \alpha_H$) and graphs are not regular (i.e ${\bf1}^{\top}L_G{\bf1}>0$ and ${\bf1}^{\top}L_H{\bf1}>0$ for normalized laplacian). This contradiction says that assumption is wrong, so one layer Chebnet's second support can distinguish 1-WL equivalent graphs whose maximum eigenvalues are not same and graphs are not regular with the same degree.
\end{proof}
Since the graph laplacians are positive semi-definite, it always yields ${\bf1}^{\top}L_G{\bf1}\geq0$ and ${\bf1}^{\top}L_H{\bf1}\geq0$ and they are zero as long as the graphs are regular with the same degree. Thus, if we add smallest positive scalar value on the diagonal of the laplacian such $L \gets L+\epsilon I$, we get rid of the necessity that graphs must be non-regular. So Chebnet become more powerful and will be able to distinguish all 1-WL equivalent regular graphs whose maximum eigenvalues are different. Considering the graph8c task, we have seen that classic ChebNet could not distinguish 44 pairs where there are 312 1-WL equivalent pairs. If we use $L \gets L+0.01 I$, the number of undistinguished pairs of graph decreased from 44 to 19, where 19 undistinguished pairs are all 1-WL equivalent and have exact the same maximum eigenvalues. On the other hand, original Chebnet was not able to distinguish 44-19=25 graphs pairs whose maximum eigenvalues are different but all of them are regular thus ${\bf1}^{\top}L{\bf1}=0$.   

\subsection{Theorem.3}
\label{section:prth3}


\begin{proof}
The number of 3-star patterns can be determined by $\sum_v \binom{d(v)}{3}$ where $d(v)$ is the degree of vertex $v$ for undirected simple graphs  \cite{pinar2017escape}. Using $f(x)=\frac{x!}{(x-3)!3!}$ as a function that operates on each element of a given vector $x$, we can calculate the number of 3-star patterns in a given adjacency matrix $A$ by ${\bf1}^{\top}f(A{\bf1})$ using operations in $\mathcal{L}_1^+$.  According to the universal approximation theory of multi layer perceptron \cite{hornik1989multilayer}, if we have enough layers, we can implement $f(.)$ as an MLP in our model. 
\end{proof}

\subsection{Theorem.4}
\label{section:prth4}


\begin{proof}
The number of triangles can be determined by using trace operator as $1/6 \times tr(A^3)$ \cite{harary1971number} which can be written by means of operations from $\mathcal{L}_2^+$.

Number of 4-cycles is determined by $1/8\times (tr(A^4)+tr(A^2)-2{\bf1}^{\top}A^2{\bf1})$ \cite{harary1971number} which can be written by means of operations from $\mathcal{L}_2^+$.
\end{proof}

\subsection{Theorem.5}
\label{section:prth5}


\begin{proof}
If $t(v)$ denotes the number triangles including vertex $v$ and $d(v)$ denotes the degree of vertex $v$, the number of tailed triangles can be found by $\sum_vt(v).(d(v)-2)$ for simple undirected graphs \cite{pinar2017escape}. Every node in a triangle has two closed walks of length 3. Thus, $t(v)=\frac{(A^3)_{v,v}}{2}$. It yields the number of tailed triangles can be found by $\frac{1}{2} \times {\bf1}^{\top}(A^3 \odot diag(A{\bf1}-2)){\bf1}$. The computation of $t(v)$ which involves the element-wise multiplication can be written with operations from $\mathcal{L}_3^+$.
\end{proof}

\subsection{Theorem.6}
\label{section:prth6}


\begin{proof}
Since the sentences in $ML(\mathcal{L}_1)$ produce a scalar value which can be reached in the graph readout layer as a sum thanks to ${\bf1}^{\top}H^{(l_{end})}$, we need to show that the MPNN can produce all possible vectors in $\mathcal{L}_1$ on the last node representation layer. Since $H^{(0)}={\bf1}$, the output of the first layer consists of linear combination of $[{\bf1}, A{\bf1}]$ because, in this case, the third term of the sum is just ${\bf1}\circ{\bf1}={\bf1}$. On the second layer, the representation consists of a linear transformation of 4 different vectors $[{\bf1}, A{\bf1},A^2{\bf1},A{\bf1}\circ A{\bf1} ]$. We can notice that these 4 vectors are the all possible vectors that $\mathcal{L}_1$ can produce up to the second level. The $diag$ operator can produce other outputs if we apply $diag(A{\bf1}).diag(A{\bf1}){\bf1}=A{\bf1}\circ A{\bf1}$. Because $diag({\bf1})=I$ cannot change anything if we use it any other expressions. Another selection would be  $A.diag(A{\bf1}){\bf1}=A^2{\bf1}$ and last option gives $diag(A{\bf1})A{\bf1}=A{\bf1}\circ A{\bf1}$. So up to $l=2$ the proof is true. Then, we follow an inductive reasoning and assume that in the $k$-th layer, Eq.\eqref{eq:ml1} produces all possible vectors ($h_1,\dots h_n$) in $\mathcal{L}_1$ and we show that it is true for $k+1$-th layer as well. In the $k+1$-th layer, the first term of the sum keeps $h_1,\dots h_n$. The second term produces $Ah_1,\dots Ah_n$. Finally, the term of the sum produces all pairs of element-wise multiplication such as $h_1\circ h_1,h_1\circ h_2, \dots h_n\circ h_n$. These are the all vectors that the language $\{.,{\bf1},diag\}$ can produce using one extra $A$ and/or $diag$ operator. The transpose operator is neglected because the adjacency matrix is symmetric. Furthermore, since at the readout layer these vectors are to be summed up, their order or the fact that they are transposed or not does not matter.

Beside, it was also shown that $diag(.)$ operator can be implemented by element-wise multiplication of vectors in \cite{geerts2020expressive} in Proposition 8.1.
\end{proof}

\subsection{Theorem.7}
\label{section:prth7}


\begin{proof}
If the given function is $\Phi(\lambda)$, it can be written by power series using the Maclaurin expansion as follows:
\begin{equation}
  \label{eq:th41}
  \Phi(\lambda)=\frac{\Phi(0)}{0!}\lambda^0 +\frac{\Phi'(0)}{1!}\lambda^1 +\frac{\Phi^{(2)}(0)}{2!}\lambda^2 +  \dots.
\end{equation} 
Thus, the frequency response can be written by power series with coefficients $\alpha_i=\frac{\Phi^{(i)}(0)}{i!}$. Using these coefficients, the convolution support can be formulated as 
\begin{equation}
  \label{eq:th42}
  C=\alpha_0UIU^{\top} +\alpha_1Udiag(\lambda)U^{\top} +\alpha_2Udiag(\lambda)^2U^{\top} +  \dots.
\end{equation} 
Since $UIU^{\top}=I=L^0$ and $Udiag(\lambda)^nU^{\top}=L^n$, we can reach the final expression:
\begin{equation}
  \label{eq:th43}
  C=\alpha_0L^0 +\alpha_1L^1 +\alpha_2L^2 +  \dots
\end{equation}

The convolution support $C$ is expressed as power series of graph laplacian $L$ as long as all order derivation of frequency response is not zero ($\Phi^{(n)}(0) \neq 0$). Since the selection of the function is based on $exp(.)$ and its derivation is never null, we can conclude that designed convolution support can be written by power series of graph Laplacian.
\end{proof}

\section{$\mathcal{L}_1$ Equivalent Graphs}
\label{app2}
\begin{figure}[ht]
\begin{center}
\centerline{\includegraphics[width=\columnwidth]{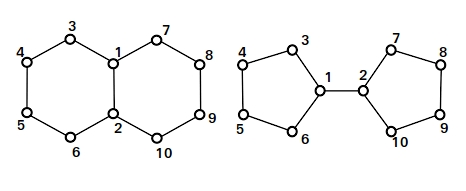}}
\caption{Decalin and Bicyclopentyl graphs are $\mathcal{L}_1$ equivalent and so 1-WL.
}
\label{dec_bicyc_enum}
\end{center}
\vskip -0.2in
\end{figure}
Figure~\ref{dec_bicyc_enum}, shows Decalin and Bicyclopentyl graphs, with a proposed node enumeration. According to these enumerations, their adjacency matrices are $A_{G}$ and $A_{H}$,  respectively
$$\scriptstyle{A_{G}}=\left ( \begin{smallmatrix} 
0   &  1  &   1  &   0   &  0   &  0  &   1  &   0  &   0 &    0\\
     1  &   0    & 0  &   0  &   0  &   1  &   0  &   0   &  0  &   1\\
     1  &   0   &  0  &   1  &   0  &   0  &   0  &   0   &  0  &   0\\
     0   &  0   &  1  &   0  &   1  &   0  &   0  &   0   &  0  &   0\\
     0   &  0   &  0  &   1  &   0  &   1  &   0  &   0   &  0  &   0\\
     0   &  1   &  0  &   0  &   1  &   0  &   0  &   0   &  0  &   0\\
     1   &  0   &  0  &   0  &   0  &   0  &   0  &   1   &  0  &   0\\
     0   &  0   &  0  &   0  &   0  &   0  &   1  &   0   &  1  &   0\\
     0   &  0   &  0  &   0  &   0  &   0  &   0  &   1   &  0  &   1\\
     0   &  1   &  0  &   0  &   0  &   0  &   0  &   0   &  1  &   0 
\end{smallmatrix} \right)
\text{~~and~~}
\scriptstyle{A_{H}}=\left ( \begin{smallmatrix}
0  &   1  &   1  &   0 &    0   &  1   &  0  &   0  &   0 &    0\\
     1  &   0  &   0  &   0   &  0  &   0  &   1   &  0   &  0  &   1\\
     1  &   0  &   0  &   1   &  0  &   0  &   0    & 0   &  0  &   0\\
     0  &   0  &   1  &   0   &  1  &   0  &   0    & 0   &  0  &   0\\
     0  &   0  &   0  &   1   &  0  &   1  &   0    & 0   &  0  &   0\\
     1  &   0  &   0  &   0   &  1  &   0  &   0    & 0   &  0  &   0\\
     0  &   1  &   0  &   0   &  0  &   0  &   0    & 1   &  0  &   0\\
     0  &   0  &   0  &   0   &  0  &   0  &   1    & 0   &  1  &   0\\
     0  &   0  &   0  &   0   &  0  &   0  &   0    & 1   &  0  &   1\\
     0  &   1  &   0  &   0   &  0  &   0  &   0    & 0   &  1  &   0\\
\end{smallmatrix}\right)$$

Their normalized Laplacian can be calculated by $L=I-D^{-1/2}AD^{-1/2}$ and gives $L_{G}$ and $L_{H}$ as follows:

$\scriptstyle{L_{G}}=\left(\begin{smallmatrix}
 1 &  -0.33  & -0.41 &        0  &       0  &       0  & -0.41  &      0   &      0   &      0\\
-0.33 &   1    &    0   &      0   &      0 &  -0.41  &       0   &      0  &       0  & -0.41\\
-0.41  &       0  &  1 &  -0.5  &      0 &        0   &      0 &        0  &       0  &       0\\
 0   &      0&   -0.5  & 1 &  -0.5  &       0  &       0 &       0     &    0    &     0\\
 0    &     0   &      0 &  -0.5 &  1 &  -0.5    &     0   &      0  &       0  &       0\\
 0  & -0.41 &        0 &        0 &  -0.5 &   1  &       0 &        0    &     0   &      0\\
-0.41    &     0  &       0  &       0  &       0   &      0&   1&   -0.5   &      0  &       0\\
 0   &      0    &     0      &   0     &    0   &      0 &  -0.5 &   1 &  -0.5  &       0\\
 0    &     0  &       0   &      0  &       0   &      0    &     0 &  -0.5 &   1 &  -0.5\\
 0 &  -0.41  &       0     &    0  &       0   &      0    &     0    &     0   &-0.5 &   1\\
\end{smallmatrix}\right)$

$\scriptstyle{L_{H}}=\left(\begin{smallmatrix}
    1  & -0.33 &  -0.41   &      0   &      0  & -0.41   &      0  &       0  &       0   &      0 \\
   -0.33   & 1   &      0    &     0   &      0  &       0  & -0.41   &      0     &    0 &  -0.41\\
   -0.41    &     0 &   1  & -0.5  &     0  &       0  &       0   &      0    &     0   &      0\\
         0   &      0 &  -0.5  &  1 & -0.5    &     0   &     0  &       0  &       0  &       0\\
         0   &      0    &     0 &  -0.5 &   1 &  -0.5   &      0   &      0    &     0  &      0\\
   -0.41  &       0   &      0   &      0 & -0.5  &  1    &     0   &      0  &       0  &       0\\
         0 &  -0.41  &       0   &      0   &      0   &      0  &  1  &-0.5  &       0   &      0\\
         0   &      0  &       0     &    0  &       0   &      0  & -0.50  &  1 &  -0.5 &       0\\
         0   &      0  &      0    &     0   &      0    &     0   &      0  & -0.50  &  1 &  -0.5\\
         0 &  -0.41  &       0 &       0   &      0    &     0   &      0   &      0 &  -0.5 &   1 \\
\end{smallmatrix}\right)$

Their second Chebnet convolution supports are $C^{(2)}_G=2/2L_G-I$ and $C^{(2)}_H=2/1.8418L_H-I$ because their maximum eigenvalues are 2.0 and 1.8418 respectively. Finally, when computing the output of the first layer by linear activation function without any learning parameters, we obtain $y_G={\bf1}^{\top}C^{(2)}_G{\bf1}=-9.9327$ and $y_H={\bf1}^{\top}C^{(2)}_H{\bf1}=-9.9269$. We observe a slight difference between these two values, which means that Chebnet can project both graphs to the different points, thus it is able to distinguish them. 

Since the maximum eigenvalues of graphs Laplacians are different, they are not cospectral as well. It means that they can also be distinguished on the basis of the number closed walks for some lengths which can be determined by trace operator. Indeed, even if  up to 4th power of the adjacency matrix, the trace operator gives the same values for both graphs, we can observe that $tr(A_G^5)=0$ whereas $tr(A_H^5)=20$. This observation is sufficient to claim that both graphs are not $\mathcal{L}_2$ equivalent.

\section{$\mathcal{L}_2$ Equivalent Graphs}
\label{app3}

\begin{figure}[ht]
\begin{center}
\centerline{\includegraphics[width=0.8\columnwidth]{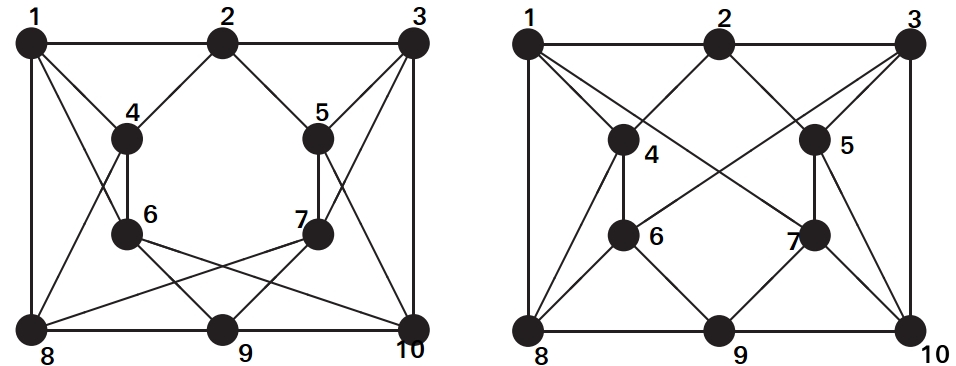}}
\caption{Cospectral and 4-regular graphs from \cite{van2003graphs} are $\mathcal{L}_2$ equivalent.
}
\label{cospec_enum}
\end{center}
\vskip -0.2in
\end{figure}

Figure~\ref{cospec_enum} shows two non-isomorphic but $\mathcal{L}_2$ equivalent graphs, where vertices are enumerated.

According to these enumerations, their adjacency matrices are the following:
$$\scriptstyle{A_{G}}=\left ( \begin{smallmatrix} 
     0  &   1  &   0  &   1  &   0  &   1  &   0  &   1 &    0  &   0\\
     1  &   0  &   1  &   1  &   1  &   0  &   0  &   0 &    0  &   0\\
     0  &   1  &   0  &   0  &   1  &   0  &   1  &   0 &    0  &   1\\
     1  &   1  &   0  &   0  &   0  &   1  &   0  &   1 &    0  &   0\\
     0  &   1  &   1  &   0  &   0  &   0  &   1  &   0 &    0  &   1\\
     1  &   0  &   0  &   1  &   0  &   0  &   0  &   0 &    1  &   1\\
     0  &   0  &   1  &   0  &   1  &   0  &   0  &   1 &    1  &   0\\
     1  &   0  &   0  &   1  &   0  &   0  &   1  &   0 &    1  &   0\\
     0  &   0  &   0  &   0  &   0  &   1  &   1  &   1 &    0  &   1\\
     0  &   0  &   1  &   0  &   1  &   1  &   0  &   0 &    1  &   0\\
\end{smallmatrix} \right)
\text{~~and~~}
\scriptstyle{A_{H}}=\left ( \begin{smallmatrix} 
     0  &   1  &   0  &   1  &   0  &   0  &   1  &   1  &   0  &   0\\
     1  &   0  &   1  &   1  &   1  &   0  &   0  &   0  &   0  &   0\\
     0  &   1  &   0  &   0  &   1  &   1  &   0  &   0  &   0  &   1\\
     1  &   1  &   0  &   0  &   0  &   1  &   0  &   1  &   0  &   0\\
     0  &   1  &   1  &   0  &   0  &   0  &   1  &   0  &   0  &   1\\
     0  &   0  &   1  &   1  &   0  &   0  &   0  &   1  &   1  &   0\\
     1  &   0  &   0  &   0  &   1  &   0  &   0  &   0  &   1  &   1\\
     1  &   0  &   0  &   1  &   0  &   1  &   0 &    0  &   1  &   0\\
     0  &   0  &   0  &   0  &   0  &   1  &   1  &   1  &   0  &   1\\
     0  &   0  &   1  &   0  &   1  &   0  &   1   &  0   &  1  &   0\\
\end{smallmatrix} \right)$$

We have seen that their normalized Laplacian eigenvalues are $\lambda_G=\lambda_H=[0, 0.44,0.61, 0.75,1.25, 1.25, 1.25, 1.25,1.56,1.64]$. Thus, they are cospectral. Considering that for cospectral graphs, the trace of any power of the adjacency matrix which gives the number of closed walks, is the same, we conclude that the trace operator does not help to distinguish these two graphs.

For instance, it can be verified that the trace of the adjacency matrix up to its 5th power is equal:
    $tr(A_G^2)=tr(A_H^2)=40$, $tr(A_G^3)=tr(A_H^3)=48$, 
$tr(A_G^4)=tr(A_H^4)=360$, and  
$tr(A_G^5)=tr(A_H^5)=920$).

However, the sentence $e(X) = {\bf1}^{\top}((X\odot X^2)^2{\bf1})^2$ which implements the element-wise multiplication from $\mathcal{L}_3$ allows to distinguish both graphs. Indeed, the computation of this sentences on $A_{G}$ and $A_{H}$ gives ${\bf1}^{\top}((A_G\odot A_G^2)^2{\bf1})^2=6032$ and ${\bf1}^{\top}((A_H\odot A_H^2)^2{\bf1})^2=5872$. Thus, these two graphs are not $\mathcal{L}_3$ equivalent (it means not 3-WL or 2-FWL equivalent as well) because the sample sentence can be explained in $\mathcal{L}_3$.

\section{$\mathcal{L}_3$ Equivalent Graphs}
\label{app4}
\begin{figure}[ht]
\begin{center}
\centerline{\includegraphics[width=\columnwidth]{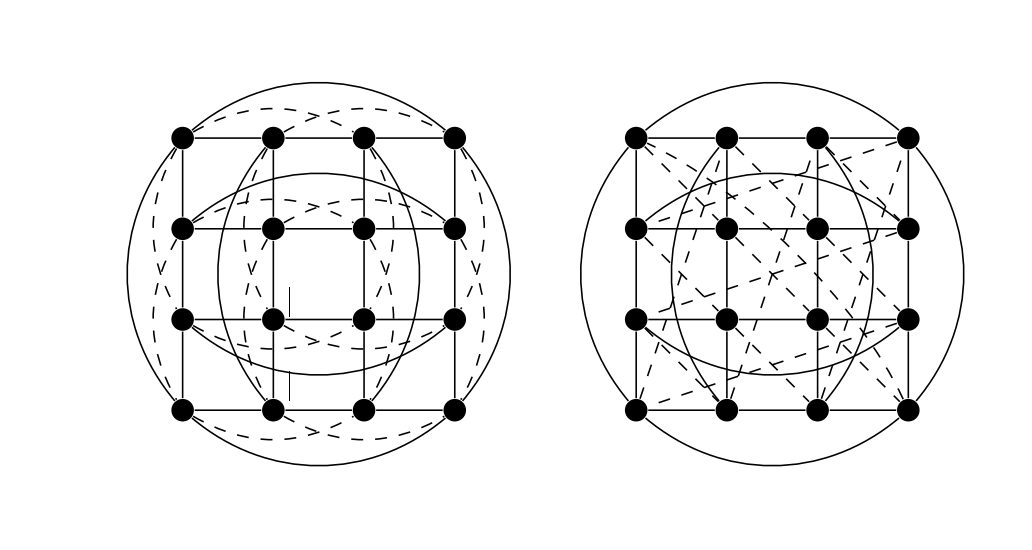}}
\caption{Strongly regular graph pair. 4 × 4-rook’s graph and the Shrikhande graph from \cite{arvind2020weisfeiler} are $\mathcal{L}_3$ equivalent.
}
\label{strongreg}
\end{center}
\vskip -0.2in
\end{figure}

Strongly regular graphs are known to be 3-WL equivalent and $\mathcal{L}_3$ equivalent as well.
Figure~\ref{strongreg} shows sample non-isomorphic graphs that are $\mathcal{L}_3$ equivalent.

When we enumerate the nodes from the top-left to the bottom-right according to their locations in the Figure~\ref{strongreg}, their adjacency matrices are the following:
$$\scriptstyle{A_{G}}=\left ( \begin{smallmatrix} 
     0  &   1  &   1  &   1  &   1  &   0  &   0  &   0  &   1  &   0  &   0  &   0  &   1  &   0  &   0  &   0\\
     1  &   0  &   1  &   1  &   0  &   1  &   0  &   0  &   0  &   1  &   0  &   0  &   0  &   1  &   0  &   0\\
     1  &   1  &   0  &   1  &   0  &   0  &   1  &   0  &   0  &   0  &   1  &   0  &   0  &   0  &   1  &   0\\
     1  &   1  &   1  &   0  &   0  &   0  &   0  &   1  &   0  &   0  &   0  &   1  &   0  &   0  &   0  &   1\\
     1  &   0  &   0  &   0  &   0  &   1  &   1  &   1  &   1  &   0  &   0  &   0  &   1  &   0  &   0  &   0\\
     0  &   1  &   0  &   0  &   1  &   0  &   1  &   1  &   0  &   1  &   0  &   0  &   0  &   1  &   0  &   0\\
     0  &   0  &   1  &   0  &   1  &   1  &   0  &   1  &   0  &   0  &   1  &   0  &   0  &   0  &   1  &   0\\
     0  &   0  &   0  &   1  &   1  &   1  &   1  &   0  &   0  &   0  &   0  &   1  &   0  &   0  &   0  &   1\\
     1  &   0  &   0  &   0  &   1  &   0  &   0  &   0  &   0  &   1  &   1  &   1  &   1  &   0  &   0  &   0\\
     0  &   1  &   0  &   0  &   0  &   1  &   0  &   0  &   1  &   0  &   1  &   1  &   0  &   1  &   0  &   0\\
     0  &   0  &   1  &   0  &   0  &   0  &   1  &   0  &   1  &   1  &   0  &   1  &   0  &   0  &   1  &   0\\
     0  &   0  &   0  &   1  &   0  &   0  &   0  &   1  &   1  &   1  &   1  &   0  &   0  &   0  &   0  &   1\\
     1  &   0  &   0  &   0  &   1  &   0  &   0  &   0  &   1  &   0  &   0  &   0  &   0  &   1  &   1  &   1\\
     0  &   1  &   0  &   0  &   0  &   1  &   0  &   0  &   0  &   1  &   0  &   0  &   1  &   0  &   1  &   1\\
     0  &   0  &   1  &   0  &   0  &   0  &   1  &   0  &   0  &   0  &   1  &   0  &   1  &   1  &   0  &   1\\
     0  &   0  &   0  &   1  &   0  &   0  &   0  &   1  &   0  &   0  &   0  &   1  &   1  &   1  &   1  &   0
\end{smallmatrix} \right)$$ 
$$\scriptstyle{A_{H}}=\left ( \begin{smallmatrix} 
     0  &   1  &   0  &   1  &   1  &   1  &   0  &   0  &   0  &   0  &   0  &   0  &   1  &   0  &   0  &   1\\
     1  &   0  &   1  &   0  &   0  &   1  &   1  &   0  &   0  &   0  &   0  &   0  &   1  &   1  &   0  &   0\\
     0  &   1  &   0  &   1  &   0  &   0  &   1  &   1  &   0  &   0  &   0  &   0  &   0  &   1  &   1  &   0\\
     1  &   0  &   1  &   0  &   1  &   0  &   0  &   1  &   0  &   0  &   0  &   0  &   0  &   0  &   1  &   1\\
     1  &   0  &   0  &   1  &   0  &   1  &   0  &   1  &   1  &   1  &   0  &   0  &   0  &   0  &   0  &   0\\
     1  &   1  &   0  &   0  &   1  &   0  &   1  &   0  &   0  &   1  &   1  &   0  &   0  &   0  &   0  &   0\\
     0  &   1  &   1  &   0  &   0  &   1  &   0  &   1  &   0  &   0  &   1  &   1  &   0  &   0  &   0  &   0\\
     0  &   0  &   1  &   1  &   1  &   0  &   1  &   0  &   1  &   0  &   0  &   1  &   0  &   0  &   0  &   0\\
     0  &   0  &   0  &   0  &   1  &   0  &   0  &   1  &   0  &   1  &   0  &   1  &   1  &   1  &   0  &   0\\
     0  &   0  &   0  &   0  &   1  &   1  &   0  &   0  &   1  &   0  &   1  &   0  &   0  &   1  &   1  &   0\\
     0  &   0  &   0  &   0  &   0  &   1  &   1  &   0  &   0  &   1  &   0  &   1  &   0  &   0  &   1  &   1\\
     0  &   0  &   0  &   0  &   0  &   0  &   1  &   1  &   1  &   0  &   1  &   0  &   1  &   0  &   0  &   1\\
     1  &   1  &   0  &   0  &   0  &   0  &   0  &   0  &   1  &   0  &   0  &   1  &   0  &   1  &   0  &   1\\
     0  &   1  &   1  &   0  &   0  &   0  &   0  &   0  &   1  &   1  &   0  &   0  &   1  &   0  &   1  &   0\\
     0  &   0  &   1  &   1  &   0  &   0  &   0  &   0  &   0  &   1  &   1  &   0  &   0  &   1  &   0  &   1\\
     1  &   0  &   0  &   1  &   0  &   0  &   0  &   0  &   0  &   0  &   1  &   1  &   1  &   0  &   1  &   0
\end{smallmatrix} \right)$$

The eigenvalues of the normalized Laplacian are equal ($\lambda_G=\lambda_H$). Both normalized Laplacians have 3 distinct eigenvalues which are 0, 0.667 and 1.33 with the respective multiplicity of 1, 6 and 9. Thus the graphs are cospectral. Since they are 3-WL equivalent, none of the sentences in $\mathcal{L}_3$ can distinguish these graphs. For instance, we have seen that ${\bf1}^{\top}((A_G\odot A_G^2)^2{\bf1})^2={\bf1}^{\top}((A_H\odot A_H^2)^2{\bf1})^2=331776$.

In order to distinguish these two graphs, we need to mimic the 3-FWL (or 4-WL) test which needs a 3-order relationship between nodes. Thus, the adjacencies will be represented by $A_G,A_H \in \mathbb{R}^{16 \times 16 \times 16}$. For any 3 nodes there are 3 different pairs and thus $2^3=8$ different states representing how these 3 nodes are connected or not. An additional state is used for the tensor diagonal. Thus, there is a total of $9$ states. The node tuple is denoted by $A_{i,j,k} \in \{0,\dots,8\}$. $0$ refers to the fact that none of three nodes are connected. $7$ refers to the fact that all nodes are mutually connected (triangle). $8$ is used for the tensor diagonal elements. We can then define an equivariant 3 dimensional tensor square operator by $(A^2)_{i,j,k}=\sum_s(A_{s,j,k}.A_{i,s,k}.A_{i,j,s})$. By summing all elements of the 3-dimensional squared adjacency where the given adjacency is for instance 0, we can distinguish these two graphs. Indeed, $\sum(A_G^2 \odot (A_G{=0}))=205632$ whereas $\sum(A_H^2 \odot (A_H{=0}))=208704$. We can then conclude that these two graphs are not 3-FWL (or 4-WL) equivalent.

\section{Result of TU Datasets}
\label{section:TUresult}

Table~\ref{TUtask} shows the results of 10-fold cross validation over studied datasets named MUTAG, ENZYMES, PROTEINS and PTC. All these datasets consist of chemical molecules where nodes refer to atoms while edges refer to atomic bonds. For these molecular datasets, node features is a one hot coding of atom types and none of the model use any edge feature even if it exists for MUTAG. In addition to these results, we also provide results on the ENZYMES dataset using extra 18-length continuous features on atoms. Using these continuous features, graph agnostic method MLP performance increases drastically from 30.8\% to 70.6\%, showing that these continuous features contain at least a part of the structural information. Models were ran for a fixed number of epochs on each fold and we select the epoch where the general accuracy is maximum on the validation set. The test procedure and train/validation split was taken from \cite{xu2018how}.

\begin{table*}
\caption{Results on TU datasets. The values are the accuracy. Edge features are not used even if they are available in the datasets. The models use a one-hot encoding of node labels as node features, while the models also use extra 18 length continuous node features for ENZYMES-cont.}
\label{TUtask}
\vskip 0.15in
\begin{center}
\begin{scriptsize}
\begin{sc}
\begin{tabular}{lcccccc}
\toprule
Model & MUTAG  & ENZYMES & ENZYMES-cont &PROTEINS & PTC \\
\midrule
MLP     & 86.6\% $\pm$ 4.95 & 30.8\% $\pm$ 4.26  & 70.6\% $\pm$ 5.22 & 74.3\% $\pm$ 4.88 & 62.9\% $\pm$ 5.89\\
GCN      & 89.1\% $\pm$ 5.81 & 49.0\% $\pm$ 4.25 & 74.2\% $\pm$ 3.26 & 75.2\% $\pm$ 5.11 & 64.3\% $\pm$ 8.35\\
GAT     & 90.1\% $\pm$ 5.84 & 54.1\% $\pm$ 5.15 & 73.7\% $\pm$ 4.47 & 75.9\% $\pm$ 4.26 & 65.7\% $\pm$ 7.97\\
GIN    & 89.4\% $\pm$ 5.60&  55.8\% $\pm$ 5.23 & 73.3\% $\pm$ 4.48 & 76.1\% $\pm$ 3.97 & 64.6\% $\pm$ 7.00 \\  
Chebnet &89.7\% $\pm$ 6.41 & 63.8\% $\pm$ 7.92 & 75.3\% $\pm$ 4.63 & 76.4\% $\pm$ 5.34 &65.5\% $\pm$ 4.94\\
PPGN    &90.2\% $\pm$ 6.62 & 55.2\% $\pm$ 5.44 & 72.9\% $\pm$ 4.18 & 77.2\% $\pm$ 4.53 & 66.2\% $\pm$ 6.54 \\
\textbf{GNNML1}   &90.0\% $\pm$ 0.42 &  54.9\% $\pm$ 5.97 &76.9\% $\pm$ 5.14 & 75.8\% $\pm$ 4.93 & 63.9\% $\pm$ 6.37 \\
\textbf{GNNML3}  & 90.9\% $\pm$ 5.46& 63.6\% $\pm$ 6.52 & 78.1\% $\pm$ 5.05 & 76.4\% $\pm$ 5.10 & 66.7\% $\pm$ 6.49\\
\bottomrule
\end{tabular}
\end{sc}
\end{scriptsize}
\end{center}
\vskip -0.1in
\end{table*}

\section{Datasets and Application Details}
\label{section:dataset}

\begin{table*}
\caption{Summary of the datasets used in our experiments.}
\label{dataset}
\begin{center}
\begin{tiny}
\begin{sc}
\begin{tabular}{l c c c c c c c c c c c c}
\toprule
  & Graph8c & SR25 & EXP & 2D-Grid& Random &Band-Pass & PROTEINS & ENZYMES & MUTAG & PTC & MNIST-75 & Zinc12K\\
\hline
Task &Iso &Iso &Iso\&2class &NReg &Reg &2class & 2class & 6class&2class &2class &10class & Reg \\
Graphs &11117 &15 &1200 &3 & 5K &5K &1113  &600 &188 &344 &70K & 12K\\
Nodes  &8.0  & 25.0   &44.44 &900.0 &18.8 & 200.0 &39.06 &32.63 & 17.93 &25.55 &  75.0  & 23.15\\
Edges &28.82 &300.0 &110.21 &3480.0 &62.67 &1072.6 &72.82 & 62.14  &39.58 & 51.92 & 694.7 & 49.83  \\
Feature &Mono &Mono &Mono &1 &Mono &1 &3label & 3label+18 &7label & 19label &1 & 21label \\

Train  &NA &NA &800 &1 &1500 &3K & 9-fold & 9-fold & 9-fold & 9-fold& 55K &10K\\
Val &NA &NA &200 &1 &1000 &1K & 1-fold   & 1-fold & 1-fold   & 1-fold & 5K &1K\\
Test  &NA  &NA &200 &1 &2500 &1K &NA &NA &NA &NA &10K &1K\\
\bottomrule
\end{tabular}
\end{sc}
\end{tiny}
\end{center}
\vskip -0.1in
\end{table*}

Table~\ref{dataset} shows the summary of the dataset used in experimental evaluation. The evaluation has been performed on four differents tasks depending on the dataset. These are graph isomorphism (Iso), graph regression (Reg), node regression (NReg) and $n$-class graph classification task (\#-Class). We did not use any edge features even if some were available. All features were defined on nodes. These features were discrete node labels coded by one-hot vectors (\#Label) and/or continuous features referred by numbers in Tab.~\ref{dataset}. We can notice that some graphs have no feature on nodes. 

We get the Graph8c and Sr25 dataset from online sources\footnote{http://users.cecs.anu.edu.au/$\sim$bdm/data/graphs.html}, EXP dataset from \cite{abboud2020surprising}, Random graph dataset from \cite{chen2020can}, 2D-Grid and Band-Pass dataset from \cite{balcilar2021analyzing}, Zinc12K from \cite{dwivedi2020benchmarking}, Mnist-75 dataset from online source\footnote{https://graphics.cs.tu-dortmund.de/fileadmin/ls7-www/misc/cvpr/mnist-superpixels.tar.gz} which was used in \cite{balcilar2021analyzing} with exactly the same procedure, PROTEINS, ENZYMES, MUTAG and PTC from TU dataset \cite{morris2020tudataset} downloaded from resources of  \cite{xu2018how}. All dataset except for EXP, Random and 2-D grid graph were used on a single task. We used EXP for graph isomorphism test and binary classification task. 2D-Grid graph was used for three different node regression tasks respectively on low-pass, band-pass and high-pass filtering effect prediction. Finally, Random graph is used on five different substructure counting tasks.  

In all cases, we used roughly 30K trainable parameters for all problems and all models. We tuned the number of layers from 2 to 5 and the number of convolution kernels in Chebnet from 3 to 5. We used Adam optimization with learning rate in $[10^{-2},10^{-3}]$ and a  weight decay in $[10^{-3},10^{-4},10^{-5}]$. We also used dropout layer before all graph convolution layers under selection of $[0, 0.1, 0.2]$ dropout rate. We used ReLU as non-linearity operation in all layers if it is not mentioned explicitly for any specific model. For classification problems, the loss function was implemented through cross-entropy. For regression problems, mean squared error was used as the loss function except on Zinc12K dataset where the loss function was mean absolute error. Unless otherwise specified, we used both sum and max readout layer after last layer of graph convolution. It is then followed by a fully connected layer which ended up with output layer.

In GNNML3, we use the eigendecomposition of normalized Laplacian to calculate the initial edge feature for all problems, except for Zinc12K and substructure counting problems where the eigen decomposition was performed on the adjacency. Each initial convolution support is set such that $\Phi_s(\lambda)=exp(-b(\lambda-f_s)^2)$, where the bandwidth parameter $b$ is set to the value of 5. The spectrum has been uniformly sampled between minimum eigenvalue and the maximum eigenvalue with a selection of $s_n=[3,5,10]$ points in order to select the band specific parameter. Thus, band specific parameter of each frequency profile can be written  $f_s=\lambda_{min}+\frac{s-1}{s_n-1}(\lambda_{max}-\lambda_{min})$ for $s \in \{1,\dots,s_n-1\}$. For the convolution support $s=0$, we used all-pass filtering named identity matrix whose frequency response is $\Phi_0(\lambda)={\bf1}$. Thus, we have a total of $s_n$ convolution supports. The 1-hop distance is always used for receptive field which corresponds to $M=A+I$. For the learning of convolution supports needed in Eq.\eqref{eq:learnsup}, we used a single layered MLP in each $mlp_{k}$ where $mlp_{1},mlp_{2},mlp_{3}: \mathbb{R}^{S} \rightarrow \mathbb{R}^{2S}$ with a sigmoid activation, and $mlp_{4}: \mathbb{R}^{4S} \rightarrow \mathbb{R}^{S}$ with ReLU activation as long as $S$ is the number of initial convolutions extracted in the preprocessing step. In Eq.\eqref{eq:ml2}, the size of the output of $mlp_5$ and $mlp_6$ is another hyperparameter where we used the same length with the first part of the Eq.\eqref{eq:ml2} defined by dimension of $W^{(l,s)}$.

Mentioned hyperparamters are optimzed for concerned model according to validation set performance if it is available. For TU dataset, since the validation and test set is not available in public split, we first created a hyperparameter tuning task by dividing the dataset one time into pre-training (80\%) and pre-validation (20\%). The optimal value  of the parameters is searched on the basis of the performance on the pre-validation set. Then, these hyperparameter values for the general test procedure as defined in \cite{xu2018how}.

Our tests were conducted with implementations of Chebnet, GCN, GIN and GAT layer provided by pytorch-geometric \cite{fey2019fast}. Besides, PPGN, GNNML1 and GNNML3 layer were implemented as a class of pytorch-geometric and our models were tested on the basis of these implementation. By doing so, we integrate the PPGN into the widely used graph library pytorch-geometric and make it publicly available beside our own proposals. 
\section{Summary of the Baseline Models}
\label{section:implementation}

\subsection{MPNN Baselines}
In this section of the appendix, we present the baseline methods which are GCN, GIN, Chebnet and GAT thanks to the general framework given by Eq.\eqref{eq:mpgnn}. Each model differs from others by selection of their convolution support $C$.

GCN uses a single convolution support given by;
\begin{equation}
  \label{eq:eqgcn}
  C=(D+I)^{-0.5}(A+I)(D+I)^{-0.5},
\end{equation}
where $D$ is the diagonal degree matrix \cite{kipf16:_semi} in Eq.\eqref{eq:mpgnn}.

Chebnet relies on the approximation of a spectral graph analysis proposed in~\citep{hammond2011wavelets}, based on the Chebyshev polynomial expansion of the scaled graph Laplacian. The number of convolution supports $C^{(k)}$ can be chosen. They are defined by \citep{defferrard16:_convol_neural_networ_graph_fast} as follows:
\begin{equation}
  \label{eq:chebeq}
  \begin{split}
  C^{(1)}=I,~~C^{(2)}=2L/\lambda_{\max}-I,\\C^{(k)}=2C^{(2)}C^{(k-1)} - C^{(k-2)}, \quad \forall k \geq 2.
  \end{split}
\end{equation}

Graph Isomorphism Network (GIN) defined in \citep{xu2018how} has a single convolution support defined as follows:
\begin{equation}
  \label{eq:gin1}
  C=A+(1+\epsilon)I,
\end{equation}
where $\epsilon$ is a parameter that makes the support trainable. Another version named GIN-0 is also defined in the same paper where $\epsilon=0$, which makes $C=A+I$. GIN proposes to use a desired number of MLP after each graph convolution. In our implementation, we use one MLP ($C=I$) after each GIN graph convolution as described in \cite{xu2018how}.

Graph attention networks (GATs) in \citep{velivckovic2017graph} proposes to transpose the attention mechanism from \cite{vaswani2017attention} into the graph world by the way of sparse attention instead of full attention in transformers. GAT convolution support can be seen as weighted, self loop added adjacency. It can be represented in Eq.\eqref{eq:mpgnn} by defining its trainable convolution supports as follows:
\begin{equation}
   \label{eq:gat}
   \left(C^{(l,s)}\right)_{v,u} = 
   \frac{e_{v,u}}{\sum_{k \in \Tilde{\mathcal{N}}(v)}e_{v,k}},
\end{equation}
where $e_{v,u}=\exp \big( \sigma(\mathbf{a}^{(l,s)} [H^{(l)}_{:v}W^{(l,s)}||H^{(l)}_{:u}W^{(l,s)}]) \big)$, 
and $\mathbf{a}^{(l,s)}$ is another trainable weight. Convolution support will be calculated from node $v$ to each element of $\Tilde{\mathcal{N}}(v)$, which shows the self-connection added neighborhood. In application of GAT, we use concatenation instead of sum in Eq.\eqref{eq:mpgnn} where the paper proposed both and there is slightly empirical advantage to use concatenation.

All MPNN baselines start with a given node features $H^{(0)}$ and provide the node representation of the next layer by Eq.\eqref{eq:mpgnn}. After the last layer, we apply a graph readout function which summarizes the learned node representation. Graph readout layer is followed by a desired number of fully connected layers ended with a number of neuron defined by targeted number of classes. 

\subsection{PPGN Baseline}
PPGN \cite{maron2019provably} starts the process with a 3-dimensional input tensor where the adjacency, edge features (if it exists) and diagonalized node features are stacked on the 3rd dimension as:
\begin{equation}
  \label{eq:eqinv}
  H^{(0)}=[A | E_1| \cdots |E_e|diag(X_1)|\cdots |diag(X_d)].
\end{equation}
Here, $X \in \mathbb{R}^{n \times d}$ gathers node features and $X_i$ is its $i$-th column vector, $E \in \mathbb{R}^{n \times n \times e}$ is edge features and $E_i \in \mathbb{R}^{n \times n}$ is its $i$-th edge feature matrix, thus initial feature tensor is $H^{(0)} \in \mathbb{R}^{n\times n\times(1+e+d)}$. 

One layer forward calculation of PPNN would be:
\begin{equation}
  \label{eq:prov1}
  H^{(l+1)}= m_3\left( \big[ m_1(H^{(l)}) \circ m_2(H^{(l)}) | H^{(l)} \big]\right)
\end{equation}
where $m_1,m_2:\mathbb{R}^{n \times n \times d_{inp}} \rightarrow \mathbb{R}^{n \times n \times d_{mid}}$ and $m_3:\mathbb{R}^{n \times n \times d_{mid}+d_{inp}} \rightarrow \mathbb{R}^{n \times n \times d_{out}}$ are trainable models that can be implemented by a one layer MLP followed by nonlinearity. $d_{inp}$ is the feature length on the 3rd dimension. $d_{mid},d_{out}$ are the feature lengths which can be seen as hyperparameters of the layer. Multiplication ($\circ$) operates between matching features and means 2d matrix multiplication for each slice which has $n \times n$ dimensions. $|$ operator is just the concatenation of two tensor on the 3rd dimension. 
The output of the model would be:
\begin{equation}
  \label{eq:prov2}
  Y= \sum_{l=1}  mlp_l\left(\sum\text{diag}(H^{(l)})~ |~ \sum\text{offdiag}(H^{(l)}) \right).
\end{equation}

We assign a function which selects the diagonal of each 2d slices of tensor as $\text{diag}: \mathbb{R}^{n \times n \times d} \rightarrow \mathbb{R}^{n \times 1 \times d}$ and function for selection the element out of the diagonal as $\text{offdiag}: \mathbb{R}^{n \times n \times d} \rightarrow \mathbb{R}^{n \times (n-1) \times d}$. We use the sum operator which performs sum over the first 2 dimensions as $\sum:\mathbb{R}^{d_1 \times d_2 \times d} \rightarrow \mathbb{R}^{d}$ and a trainable model that may be implemented by an MLP $mlp_l:\mathbb{R}^{2d} \rightarrow \mathbb{R}^{d_{y}}$, transforms the given vector into the targeted output representation length.  

The one can see that in each layer, PPNN keeps $H^{(l)} \in \mathbb{R}^{n\times n\times d_l}$, thus its memory usage is in $\mathcal{O}(n^2)$. Since there is a matrix multiplication in Eq.\eqref{eq:prov1}, its computation complexity is in $\mathcal{O}(n^3)$ when using the naive matrix multiplication operations. The PPNN paper mentioned that the computational complexity can be decreased by using effective matrix multiplication, but it is the same for all algorithms as well. For this reason, we think that taking the naive implementation into account makes more sense to do a fair comparison. In addition, again because of matrix multiplication, its update mechanism is not local. Because of calculation of the $u,v$ node pairs representation in Eq.\eqref{eq:prov1}, it needs to perform $\sum_k H^{(l)}_{u,k}.H^{(l)}_{k,v}$. That means that for each pair of nodes, $k$ should be all nodes in the graph regardless how far away the node $k$ from the concerned nodes $u,v$. In other words, very far away nodes feature affect the concerned node.    

Even though PPNN \cite{maron2019provably} is a  very straight forward algorithm and has provable 3-WL power, the experimental results reported in the papers are not at the state of the art \cite{maron2019provably,dwivedi2020benchmarking}. We believe that this can be at least partly explained by some implementation problems. Indeed, it was implemented by gathering same size graphs into batches in order to handle graphs of different size in a dataset. So the batches do not consist of randomly selected graphs in each epoch during the training phase. In our implementation, we first find the maximum size of the graph denoted as $n_{max}$. Then, we create an initial tensor in Eq.\eqref{eq:eqinv} in dimension of $\mathbb{R}^{n_{max} \times n_{max} \times 1+e+d}$ where left top $n \times n \times 1+e+d$ part of the tensor is valid, and the rest is  zero. We also keep the valid part of the tensor diagonal and out of diagonal part mask in $M_0,M_1 \in \{0,1\}^{n_{max} \times n_{max}}$ that shows which element is valid in the diagonal and which element is valid out of the diagonal of the representation tensor. Since some part of the tensor $H^{(l)}$ are not valid, we need to prevent to assign value after application of trainable model $m_k$ in Eq.\eqref{eq:prov1}, because it affects the matrix multiplication result. One solution may be to mask the MLP result by $M_0+M_1$. Finally, we implement Eq.\eqref{eq:prov2} by selection diagonal and off-diagonal element by previously prepared mask matrices by $\sum\text{diag}(H^{(l)})=\sum M_0 \odot H^{(l)}$ and  $\sum\text{offdiag}(H^{(l)})=\sum M_1 \odot H^{(l)}$. By doing so, we can put any graph into same batch. These principles have been implemented as a class of the widely used open-source pytorch geometric library.

\end{document}